\documentclass{ifacconf}
\usepackage{natbib}
\usepackage{graphicx} 
\usepackage{epsfig} 
\usepackage{epstopdf}
\usepackage{times} 
\usepackage{amsmath}
\usepackage{amsfonts}
\usepackage{amssymb}
\usepackage{subfigure}
\usepackage{color}
\usepackage[utf8]{inputenc}
\usepackage[T1]{fontenc}
\usepackage{tikz}
\usetikzlibrary{shapes,arrows,positioning,fit}
\graphicspath{{./figs/}}
\newtheorem{problem}{Problem}
\newtheorem{lemma}{Lemma}
\newtheorem{assumption}{Assumption}
\newtheorem{theorem}{Theorem}
\newtheorem{corollary}{Corollary}

\newcommand{\integernonnegative}{\ensuremath{\mathbb{Z}}_{\ge 0}}
\newcommand{\real}{\ensuremath{\mathbb{R}}}

\newcommand{\until}[1]{\{1,\dots, #1\}}
\newcommand{\subscr}[2]{#1_{\textup{#2}}}
\newcommand{\supscr}[2]{#1^{\textup{#2}}}

\newcommand{\diag}[1]{\operatorname{diag}(#1)}

\newcommand{\scale}{\operatorname{s}}
\newcommand{\arm}{\operatorname{arm}}
\newcommand{\cent}{\operatorname{c}}

\newcommand{\vect}[1]{\boldsymbol{#1}}
\newcommand{\ones}[1][]{\vect{1}_{#1}}

\newcommand{\neighbor}[1]{\mathcal{N}_{#1}}
\newcommand{\agentcen}[1]{c_{#1}}
\newcommand{\estcen}{c}
\newcommand{\area}{\operatorname{area}}

\newcommand{\longthmtitle}[1]{\mbox{}{\text{(#1).}}}

\newcommand\oprocendsymbol{\hbox{$\bullet$}}
\newcommand\oprocend{\relax\ifmmode\else\unskip\hfill\fi\oprocendsymbol}

\begin{document}

\tikzstyle{block} = [rectangle, draw, fill=blue!20, 
    text width=5em, text centered, rounded corners, minimum height=2.5em]
\tikzstyle{bigblock} = [rectangle, draw, fill=blue!20, 
    text width=7.5em, text centered, rounded corners, minimum height=2.5em]    
\tikzstyle{line} = [draw, -latex']
\tikzstyle{cloud} = [draw, ellipse,fill=red!20, node distance=3cm,
    minimum height=2em]

\begin{frontmatter}

    
    \title{Gesture based Human-Swarm Interactions  for Formation
    Control using interpreters}


\author[First]{Aamodh Suresh} \author[Second]{Sonia Mart{\'\i}nez}

\address[First]{Department of Mechanical and Aerospace Engineering,
  University of California at San Diego, La Jolla, CA 92093, USA
  (e-mail: aasuresh@eng.ucsd.edu).}  \address[Second]{Department of
  Mechanical and Aerospace Engineering, University of California at
  San Diego, La Jolla, CA 92093, USA (e-mail: soniamd@eng.ucsd.edu).}

\begin{abstract}                
  We propose a novel Human-Swarm Interaction (HSI) framework which
  enables the user to control a swarm shape and formation. The user
  commands the swarm utilizing just arm gestures and motions which are
  recorded by an off-the-shelf wearable armband. We propose a novel
  interpreter system, which acts as an intermediary between the user
  and the swarm to simplify the user's role in the interaction. The
  interpreter takes in a high level input drawn using gestures by the
  user, and translates it into low level swarm control commands. This
  interpreter employs machine learning, Kalman filtering and optimal
  control techniques to translate the user input into swarm control
  parameters. A notion of Human Interpretable dynamics is introduced,
  which is used by the interpreter for planning as well as to provide
  feedback to the user. The dynamics of the swarm are controlled using
  a novel decentralized formation controller based on distributed
  linear iterations and dynamic average consensus. The framework is
  demonstrated theoretically as well as experimentally in a 2D
  environment, with a human controlling a swarm of simulated
  robots in real time.

\end{abstract}


\end{frontmatter}

\section{Introduction}
\begin{figure}
\centering
\begin{tikzpicture}[node distance = 1.7cm, auto]
    \node [block] (user) {Human User};
    \node [block, right =1cm of user] (armband) {Wearable Device};
    \node [block, right =1cm of armband] (decoder) {Intention Decoder};
    \node [block, below of= decoder] (interpreter) {Planner\\};
    \node [bigblock, below of=interpreter] (controller) { Decentralized Swarm Controller};
    \node [block, left =1cm of controller] (swarm) {Robot Swarm};
    \node[draw,densely dashed,fit=(decoder) (interpreter)] (BigBox) {};
	\node[left =0.01cm of BigBox] {Interpreter};    
    
    \path [line] (user) -- node [midway,above] {$v^h$} (armband);
    \path [line] (armband) -- node [midway,above] {$o$} (decoder);
    \path [line] (decoder) -- node [midway,right]{$\hat{v}$} (interpreter);
    \path [line] (interpreter.215) -- node [midway,left]{$v^s$} (controller.145);
    \path [line] (interpreter) -- node[midway,below]{$y$} ++(-3,0) -| (user);
    \path [line] (controller.35) -- node [midway,right] {$p_0$} (interpreter.325);
    \path [line] ([yshift=0.2cm]controller.west) -- node[midway,above]{$v_i$} ([yshift=0.2cm]swarm.east);
    \path [line] ([yshift=-0.2cm]swarm.east) -- node[midway,below]{$x_i$} ([yshift=-0.2cm]controller.west);
\end{tikzpicture}\quad
\caption{Workflow of proposed Human Swarm Interface with wearable. The
  user communicates their intent $v^h$ through the Myo armband which
  produces observations $o=(\supscr{o}{emg},\supscr{o}{imu})$
  according to Section~\ref{sec:framework}. The decoder estimates the
  user intent $\hat{v}$ from observations $o$. The planner uses
  $\hat{v}$ to optimally plan a set of intermediate goals denoting the
  interpreter's command $v^s$. The decentralized controller present in
  each agent $i$ then tries to reach the $v^s_i$ by computing the
  velocities $v_i$.
} \label{fig:ASIworkflow}
\end{figure}
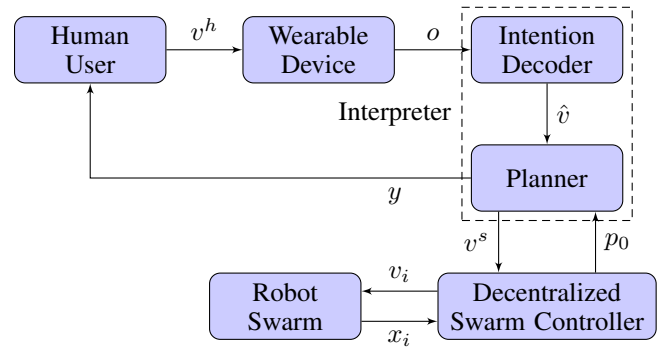

\textit{Motivation.} Due to recent advances in technology, 
the field of swarm robotics has
become pervasive in the research community while slowly permeating to the
industry. Although the coordination of multiple robots such as
foraging, coverage, and
flocking~(\cite{ROS-JAF-RMM:06,AJ-JL-ASM:03,FB-JC-SM:09})
 has
received much attention, the human interaction with robotic swarms is
less understood(~\cite{AK-WP-NC-KS-ML:16}).  Thus, according to the
latest Robotics Roadmap\footnote{\begin{tiny} Christensen, H. I., et
    al. "A roadmap for US robotics: from internet to robotics."
    (2016). \texttt{http://jacobsschool.ucsd.edu/contextualrobotics/docs/rm3-final-rs.pdf}
\end{tiny}} a top priority in
swarm robotics is the development of unifying HSI frameworks, the
elucidation of rich set of HSI examples, and their comparison. In
particular, there is a need to develop novel intuitive interfaces for
humans to communicate their intentions to swarms and make it easier
for humans to interpret swarms. At the same time, a swarm may require
high dimensional and complex control inputs which cannot be
intuitively given by a human. 
Motivated by this, we propose to build a novel supervisory interpreter
(Figure~\ref{fig:ASIworkflow}) to bridge the human and the swarm, which
is essential to ensure the effectiveness of a HSI system. We consider
the particular problem of formation control, where the human can
intuitively draw shapes in the air with his/her arm, which is
translated into an effective distributed controller.

\textit{Related Work.} According to recent surveys on HSI~(\cite{AK-WP-NC-KS-ML:16}) and human multi-agent systems~(\cite{AF:16}), humans either take a supervisory(~\cite{KS-EF:12}), direct(~\cite{TS-AF-HK-ME:15}), shared~(\cite{AF-CS-MR-HHB-PRG:12}) or environmental(~\cite{ZW-MS:16}) control role in an HSI framework. Our architecture however, allows humans to provide high level supervisory inputs that are also direct and detailed at the same time, thus allowing a high degree of control with lesser human effort for large swarms. Most of the HSI frameworks design have been human-centric and focused on direct control of swarms either through teleoperation or proximal interaction; see
e.g.~\cite{JN-AG-LMG-GAC:14,TS-AF-HK-ME:15}. Due to complicated swarm dynamics, the human will quickly be overwhelmed and would not make the best decisions, as in our previous work~\cite{AS:16,AS-MS:16}. Our planner addresses this by generating an intuitive human-approved swarm-friendly plan for the swarm to follow. 
More recently, gesture based techniques
along with speech, vision and motion have been used together to
interact with small teams of robots in \cite{JAM-SHL-PL-RS-PB:15} and
\cite{BG-LMG-GAC:16}. These works rely on proximal multi-modal interaction
schemes which require complex hardware setup to interpret the human
gestures, which is not practical for large scale swarms. We rely on a single wearable device without any other external electronics, which makes the implementation more practical. With respect to formation control for large scale swarms~\cite{MR-AC-RN:14} and~\cite{JAM-AB-MF-RS-PB:12} have only used predefined shapes and images as inputs for the swarm, which facilitates only supervisory control for a HSI system. But in our approach the swarm is capable of understanding intuitive human intention with the aid of the interpreter.   
 

\textit{Statement of Contributions.} We propose a novel HSI framework
where we consider both a human agent and a dynamic swarm, with an
interpreter acting as an bridge between the two. By means of it, the
user can communicate their intentions intuitively and naturally,
without having an in depth understanding of the swarm dynamics. At the
same time, the swarm receives control subgoals in their domain and
need not spend resources to decode the user's intention. The paper
presents contributions in the following three aspects. On the
human-interpreter interaction side, we formulate a novel intention
decoder using Kalman Filtering and HMMs for simultaneous dynamic and
static gesture decoding utilizing the IMU and EMG sensors,
respectively. This method increases intuitiveness as preliminary tests
have suggested that the human quickly learns to adapt to this
interface, with results being comparable to a standard interfaces like
a computer mouse. Second, we further exploit the interpreter element
to devise control subgoals that are efficient for the swarm, and which
require global information that is not easily accessible for the
swarm. In this way, the interpreter solves a planning problem with the
goal of controlling the swarm efficiently while following an intuitive
behavior. Third, we present a novel discrete second-order distributed
formation controller for the swarm that combines the Jacobi
Overrelaxation Algorithm and dynamic consensus to guarantee the
convergence of a (second-order integrator) swarm to a desired shape,
scaling, rotation and displacement. Our controller relies only on the
position information of each agent and communication with their
neighbors using variable communication radii, which provides a
practical setting.  Finally, we highlight a contribution on the
integration of diverse tools from control theory, network science,
machine learning, signal processing, optimization and robotics that
serve to articulate our HSI framework.

\textit{Paper
  Organization.} Section~\ref{sec:Prelim} presents preliminary
concepts required to build our framework in
Section~\ref{sec:framework}, which includes the problem statement.
We then describe our approach taken to solve each aspect of the
problem statement in Section~\ref{sec:technical_approach}. Next, we
state and discuss our results using our proposed approach in
Section~\ref{sec:results}. We finally present conclusions in Section~\ref{sec:conclusion}.
\section{Preliminary Concepts}
\label{sec:Prelim}
This section introduces the basic notation and concepts used to
construct our HSI framework.
\subsection{Basic Notations} 
\label{sec:notations}
We let $\real$ denote the space of real numbers, and
$\integernonnegative$ the space of positive integers. Also, $\real^n$
and $\real^{M \times n}$ denote the $n$-dimensional real vector space
and the space of $M\times n$ real matrices, respectively. We use
$\mathbb{P}$ to denote the set of $n$ dimensional polygonal shapes. 
   In what follows,
$\ones[M] \in \real^M$ are column vector of ones, $\mathbf{I} \in
\real^{M\times M}$ is the identity matrix, and $\mathbf{O} \in
\real^{M\times n}$ denotes a matrix of zeros. In what follows, $\|.\|$
denotes the Euclidean norm. Given a matrix $A \in \real^{M \times M}$,
its eigenvalues are denoted by $\{\lambda_1^A,\dots, \lambda_M^A\}$,
enumerated by their increasing real parts. The $\supscr{i}{th}$ row of
a matrix $A$ is denoted by $A_{i}$.


\subsection{Graph Theory Notions}
\label{sec:graph_theory}
Here, we introduce some basic Graph Theory notations which
will be used in the sequel.  Readers can refer
\cite{FB-JC-SM:09,CDG-GFR:01} for more details on Graph Theory and
its application to robotics. 

Consider a swarm of $M$ agents in $\real^n$. Let $p_i(t),v_i(t) \in
\real^n$ denote the position and velocity respectively of the
$\supscr{i}{th}$ agent at time $t$.  We denote by $p(t) \in~
\real^{M\times n}$ the position of the whole swarm defined by
$p(t)=~[p_1(t)^\top, \dots , p_M(t)^\top]^\top$. 

We model the communication among agents by means of an undirected
$\nu$-disk communication graph $\mathcal{G}_{\nu} =(V,E_\nu(p))$,
where $V=\until{M}$ denotes the set of agents (vertices of the graph),
and $E_{\nu}(p) \subset V \times V$, denotes the set of edges.  In
particular, $(i,j) \in E_\nu(p)$ if and only if $\|p_i - p_j\|\le
\nu$.  The entries of the associated adjacency matrix $A(p)\in
\real^{M\times M}$ become:
\begin{equation*}
  a_{ij}=
  \begin{cases}
    1, & \mbox{ if }\ \|p_i-p_j\| \leq \nu ,\\
    0, & \mbox{ otherwise}.
  \end{cases} 
\end{equation*}
The neighbor set $\neighbor{i}$ for the $\supscr{i}{th}$ agent is
given by $\neighbor{i}:=\{j\ |\ a_{ij}=1\}$.  Associated with
$\mathcal{G}_\nu$, we consider a weight-balanced
weighting $W(t) \in
\real^{M\times M}$, where $W(t)$ is the metropolis weight matrix corresponding to
the communication graph $\mathcal{G}_\nu$; see~\cite{LX-SB:04}.
 With $d_i=A_i(p)\ones[M]^{\top}$ 
  being
the out degree of the $\supscr{i}{th}$ agent,
$W$ is given by:
\begin{equation}
  w_{ij}=
  \begin{cases}
    1/(1+\max\{ d_i,d_j\}), & \mbox{if } (i,j) \in E_\nu(t), \\
    1 - \sum_{k\in \neighbor{i}}(1/(1+\max\{ d_i,d_j\})),&\mbox{if } i=j, \\
    0, &  \mbox{otherwise}.
  \end{cases}
  \label{eqn:weight matrix}
\end{equation} 
Since we consider an undirected graph the matrix $W$ is symmetric and
doubly stochastic.  From equation~\eqref{eqn:weight matrix} the graph
$\mathcal{G}_{w}$ is balanced as
$\ones[M]W=W\ones[M]^{\top}=\ones[M]$. We denote by $D \in
\real^{M\times M}$ the diagonal degree matrix of $\mathcal{G}$ with
$d_{i}$, the degree of node $i$, being the $\supscr{i}{th}$ diagonal
entry of $D$. The Laplacian matrix $L \in \real^{M \times M} $ of the
graph $\mathcal{G}_\nu$ is given by $L=D-A$, and the normalized laplacian matrix is given by $L^N=D^{\frac{-1}{2}}LD^{\frac{-1}{2}}$ . Similarly the weighted
Laplacian matrix is given by $L^W=\mathbf{I}-W$. The connectivity
properties of a graph are captured by the second smallest eigenvalue
$\lambda_2$ of the Laplacian matrix $L$. We can also express
connectivity in terms of $\lambda_2^W$ and $\lambda_2^N$. We can say
that the respective graph is connected if $\lambda_2^W,\lambda_2^N >
0$, and connectivity increases with increase in
$\lambda_2^W,\lambda_2^N$. 

\section{Proposed Framework and Problem Formulation}
\label{sec:framework}

Here, we first introduce the various timescales involved in the
interactions, and propose a new HSI framework, while providing a
description of its components.  Later, we identify the various
problems  to be solved to implement this framework.

\textit{Timescales Involved.} We assume that the interactions between
the human, interpreter and the swarm, and the dynamic update of the
swarm, may occur at time scales that go from coarser to finer
resolution. In this way, human and interpreter may interact at
discrete times that are a multiple of $\subscr{\tau}{h}$, the
interpreter and the swarm may interact at multiples of
$\subscr{\tau}{int} < \subscr{\tau}{h}$, while the swarm dynamic
update times occur at multiples of $\subscr{\tau}{s} <
\subscr{\tau}{int} $.  In what follows, we identify $T \equiv T
\subscr{\tau}{h}\ge 0$ (resp.~$l \equiv l \subscr{\tau}{int}$, and $t
\equiv l \subscr{\tau}{s}$) and we distinguish these integers as
belonging to $T\in \supscr{\integernonnegative}{h} \equiv
\integernonnegative$ (resp.~$l \in \supscr{\integernonnegative}{int}
\equiv \integernonnegative$, and $t \in
\supscr{\integernonnegative}{s} \equiv \integernonnegative$.)  We use
the time variable $t$ for the wearable device as it operates at a fast
rate, similar to the swarm.

\textit{Proposed Framework.} The user specifies their intentions which
are translated by the interpreter and in turn communicated to the
swarm.  The human uses a wearable device called the MYO
armband\footnote{https://www.myo.com/} which observes the human
intended swarm command.  By means of it, the user specifies a desired
formation shape $S \in \mathbb{P}$, 
centroid $\cent \in \real^2$, orientation $\theta \in \real$, and
scaling $\scale \in \real$ for the swarm. These parameters make up the
desired human intention $v$ which the interpreter decodes as
$\hat{v}$, where $v,
\hat{v}~:~\supscr{\integernonnegative}{int}~\rightarrow~\mathbb{P}~\times~\real^2
\times~\real~\times~\real $.
  The MYO armband receives the human intention $v(T)$
as Electromyography (EMG) signals $\supscr{o}{emg}(\tau)$ and Inertial
Measurement Unit IMU signals $\supscr{o}{imu}(\tau)$, where $\tau \in
[(T-~1)\subscr{\tau}{h},T\subscr{\tau}{h}]$.

The interpreter first uses a decoder
(Section~\ref{sec_gesture_decoding}) to translate human intentions
$v(T)$ into $\hat{v}(T)$. Then it translates $S(T)$ in $\hat{v}(T)$ to
desired relative agent positions $z^f(T) \in \real^{M\times n}$ which
best depicts the swarm shape. The swarm also has an operation mode
$\mu(t) \in \until{m}$ corresponding to $m$ different communication
ranges for each agent of the swarm. We have the notion of swarm
operating cost involving $\mu(t)$ as a trade-off between network
connectivity and network maintenance costs. We also introduce the
notion of Human Interpretable Dynamics (HID), which represents easily
understandable swarm dynamics by the Human. Both these concepts will
be elucidated in Section~\ref{sec:planner}.

Now, Given a desired formation $z^f(T)$ and the current state $p(0)$,
the interpreter then determines the set of switching intermediate
goals $V^s = \{v^s(1),...,v^s(N)\}$ with $v^s(l)
=\{z(l),\scale(l),\cent(l),\theta(l),\mu(l) \}$, $l\in \until{N}$ and
$N$ being the time horizon for switching. These intermediate goals
$V^s$ follow the HID and are optimal with respect to the swarm
operating costs. These intermediate goals represent way points and
intermediate shapes which will be communicated to the swarm. These
parameters constitute the high-level commands that the swarm receives
and executes via a distributed algorithm.  That is, our swarm employs
a decentralized control scheme detailed in
Section~\ref{sec:swarm_controller} to reach $v^s(l)$.
Figure~\ref{fig:ASIworkflow} illustrates the work-flow of our proposed
framework. Thus, from here, we need to solve the following
problems to complete our framework:

\begin{problem}\longthmtitle{Human Intention Decoder}
\label{prob_decoder}
Given the observations $\supscr{o}{imu}(t)$ and $\supscr{o}{emg}(t)$
from the Myo armband, design a decoder to get the desired human
intention $\hat{v}(T)$. 
\end{problem}

\begin{problem}\longthmtitle{Behavior Specifier}
\label{prob_behavior}
Given the desired human intention $\hat{v}$,
design an algorithm to produce the goal behavior $V^s$ which can
be understood by the swarm.
\end{problem}

\begin{problem}\longthmtitle{Planning Algorithm}
\label{prob_switching}
Given the goal behavior $V^s(T)$, generate the
set of optimal intermediate behavior subgoals $\{v^s(l)\}$ with $l \in
\until{N} \cap \supscr{\integernonnegative}{int}$, and $N$ denoting
the time horizon, and $N \subscr{\tau}{int} \le T \subscr{\tau}{h}$
which follow human-interpretable dynamics and minimize swarm operating
costs.
\end{problem}

\begin{problem}\longthmtitle{Distributed Swarm Controller}
\label{prob_controller}
Given the command $v^s(l)$, for some $l \in
\supscr{\integernonnegative}{int}$, design a distributed algorithm to
drive the swarm to the intermediate shape $z(l)$ with scaling
$\scale(l)$, rotation $\theta(l)$ and centroid $\cent(l)$ using
operation mode $\mu(l)$ from some initial position $p(l-1)$.
\end{problem}

\begin{problem}\longthmtitle{User Interface Design and Feedback}
\label{prob_feedback}
Develop a Graphical user interface (GUI) for the human to communicate
their intention $v$ to the interpreter and receive feedback about the
decoded intention $\hat{v}$ and the state of the swarm.
\end{problem}
We propose solutions to the above problems in
Section~\ref{sec:technical_approach}.

\section{Technical Approach}
\label{sec:technical_approach}
The following subsections describe the proposed solutions to the
problems of Section~\ref{sec:framework}. 
\subsection{Problem 1:Intention Decoding}
\label{sec_gesture_decoding}
\begin{figure}
\centering
{\includegraphics[width=0.95\linewidth]{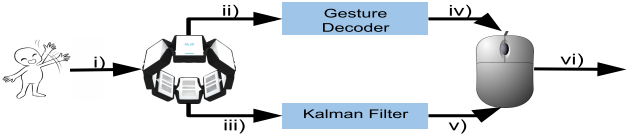}
  \caption{The user intention decoder system. i) The user conveys
    their intention $v(t)$ through arm movement and gestures. ii) The
    Myo armband captures the gestures as EMG signals
    $\supscr{o}{emg}(t)$ which are read by the gesture decoder. iii)
    Arm movements are captured as IMU signals $\supscr{o}{imu}(t)$ and
    sent to a Kalman filter. iv) The HMM based decoder provides gestures
    which are mapped to mouse clicks and scrolls. v) The updated state of
    the Kalman filter is used to assign mouse position. vi) Shape $S$
    and centroid $\cent$ are specified using the GUI
    (Figure. \ref{fig:gui}) using iv) and v)}
\label{fig:myo_loop}}
\end{figure}

\begin{figure}
 \centering
 \subfigure[\scriptsize{Fist}]
{\includegraphics[width=0.09\textwidth, height=2.5cm]{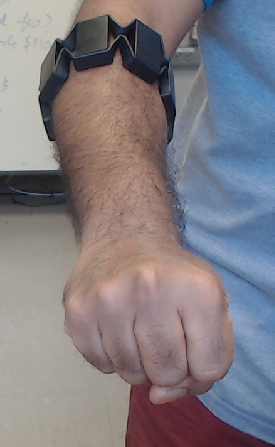}}
\subfigure[\scriptsize{Spread}]
{\includegraphics[width=0.09\textwidth, height=2.5cm]{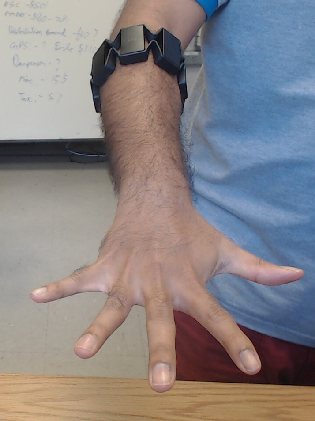}}
\subfigure[\scriptsize{Wave Up}]
{\includegraphics[width=0.09\textwidth, height=2.5cm]{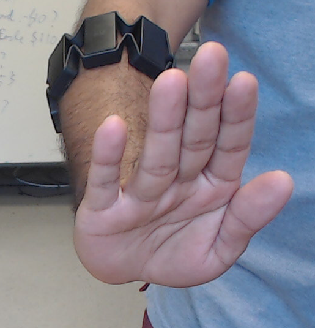}}
\subfigure[\scriptsize{Wave down}]
{\includegraphics[width=0.09\textwidth, height=2.5cm]{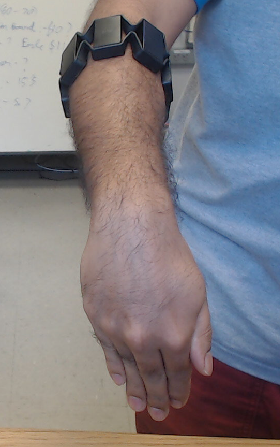}}
\subfigure[\scriptsize{Normal}]
{\includegraphics[width=0.09\textwidth, height=2.5cm]{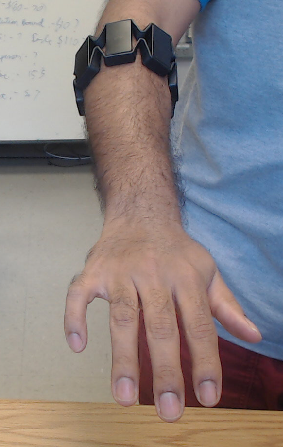}}
\subfigure[\scriptsize{Left click}]
{\includegraphics[width=0.09\textwidth, height=2.5cm]{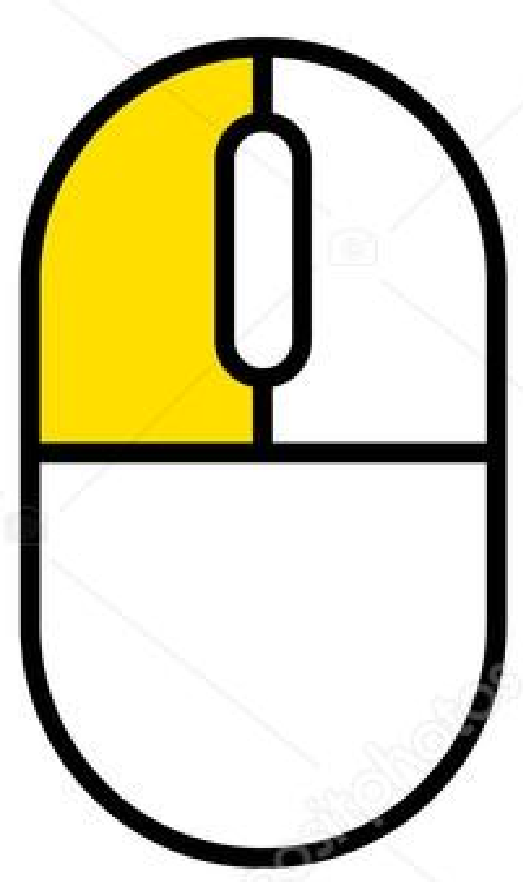}}
\subfigure[\scriptsize{Right click}]
{\includegraphics[width=0.09\textwidth, height=2.5cm]{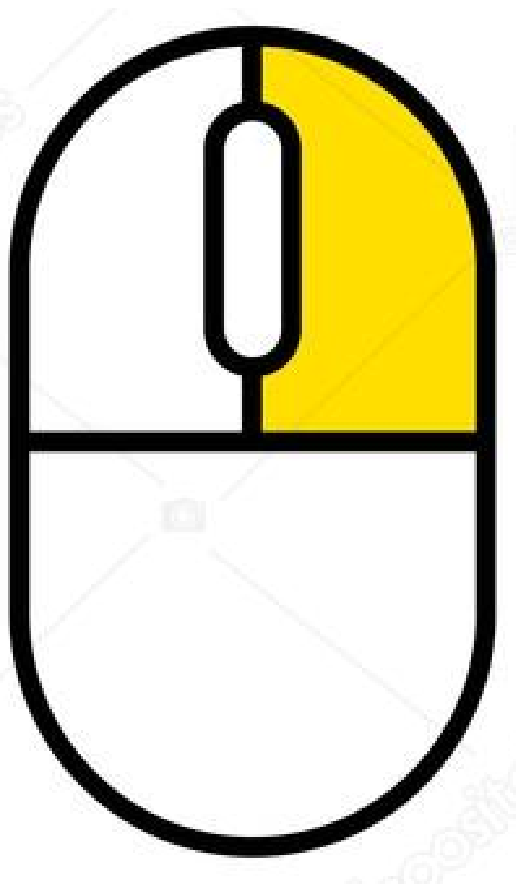}}
\subfigure[\scriptsize{Scroll up}]
{\includegraphics[width=0.09\textwidth, height=2.5cm]{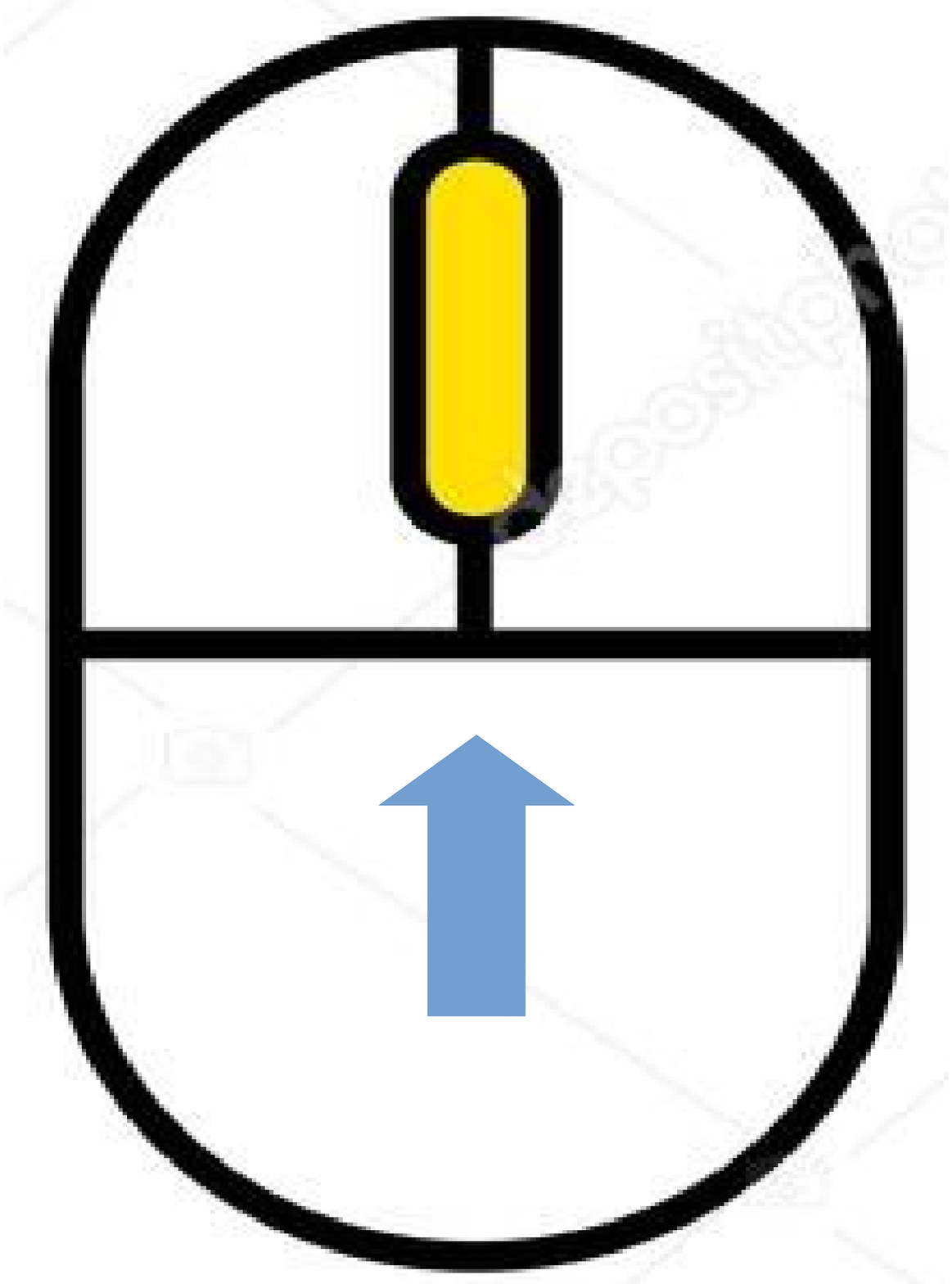}}
\subfigure[\scriptsize{Scroll down}]
{\includegraphics[width=0.09\textwidth, height=2.5cm]{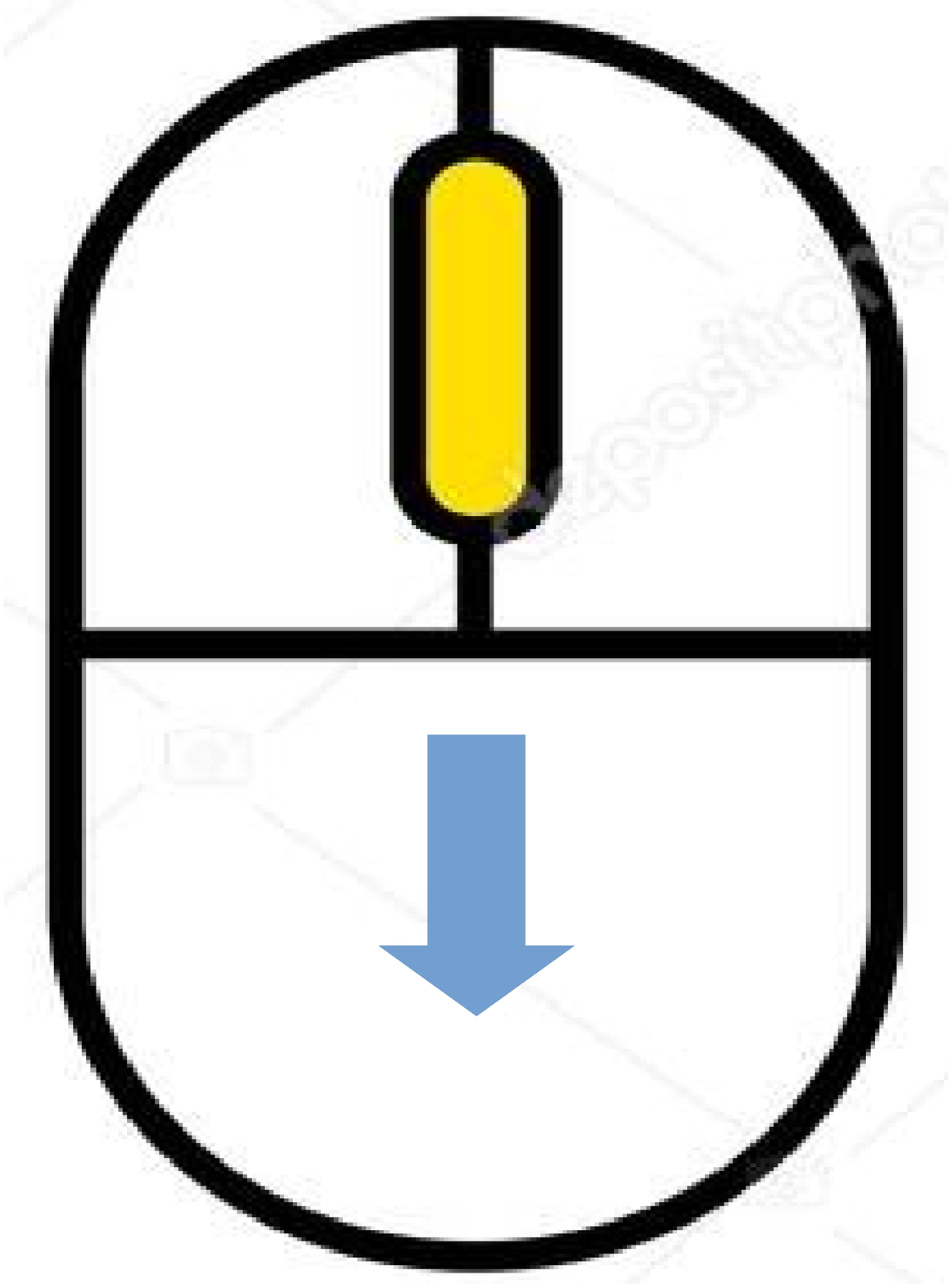}}
\subfigure[\scriptsize{Normal}]
{\includegraphics[width=0.09\textwidth, height=2.5cm]{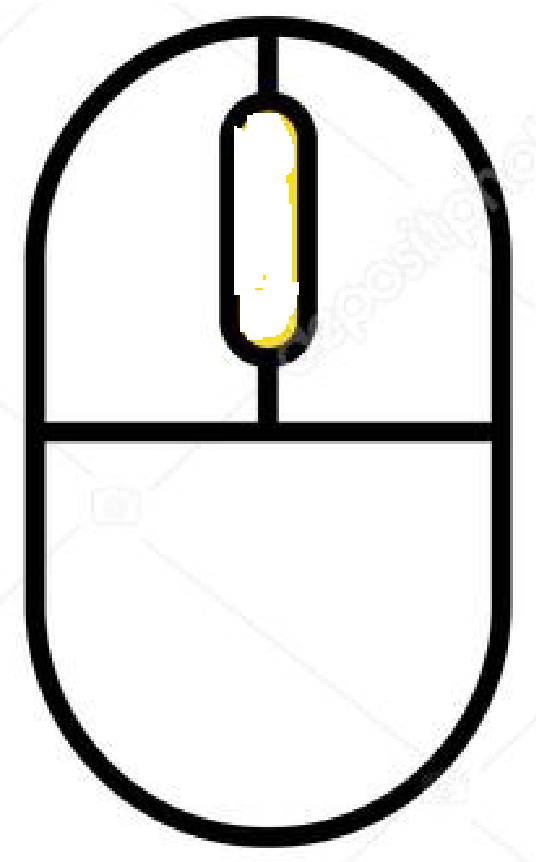}}
\caption{(a)-(e) show the various gestures used and (f)-(j) indicate
  the corresponding mouse
  functionalities. \label{fig:gesture_mapping}}
\end{figure}

The user conveys their intention $v$ through gestures and arm movement
which are recorded by the Myo armband as EMG signals. There are $8$
spatial EMG sensors on the Myo armband which generate EMG signals
$\supscr{o}{emg}(t) \in \real^8$ at every time $t$.  
The 9 DoF IMU provides 3D acceleration, 3D angular velocity, and 3D
angular orientation values. We only consider the planar angular
velocity and orientation signals and hence the relevant IMU signals
$\supscr{o}{imu}(t) \in \real^4$ at time $t$ are used. The first
aspect of the intention decoder is to decipher discrete gestures
$\supscr{o}{gs}(t) \in \{0,1,2,3,4\}$ from EMG signals
$\supscr{o}{emg}(t)$. Then it deciphers the state of the arm $\arm(t)
\in \real^4$ 
   consisting of planar arm position $\arm^p(t) \in
\real^2$ and planar arm velocity $\arm^v(t) \in \real^2$ from IMU
signals $\supscr{o}{imu}(t)$. The gestures $\supscr{o}{gs}$ and arm
state $\arm$ are translated to mouse movement and mouse clicks, which
provide feedback of the decoded intended gesture $\hat{v}$ to the
user. This pipeline is described in Figure~\ref{fig:myo_loop}.
We use a custom Hidden Markov Model (HMM) based approach to decode the
gestures $\supscr{o}{gs}(t)$ from $\supscr{o}{emg}(t)$.  In this work,
we introduce the use of five gestures and map them to mouse functions as
shown in Figure~\ref{fig:gesture_mapping}. 
   We implement a Kalman filter based movement
decoder which uses the gyroscope and magnetometer signals
$\supscr{o}{imu}(t)$ from the IMU of the Myo armband and maps it to
arm state $\arm$. The next few paragraphs give an insight about our
proposed intention decoder, however the complete details of this
pipeline are omitted due to space constraints. 

\subsubsection{Gesture decoding using HMM}
We use HMM, see~\cite{LRR:89}, a common probabilistic machine learning
technique to decode gestures from the EMG signals. Our HMM
implementation uses discrete states which are the gestures
$\supscr{o}{gs}(t) \in \{0,1,2,3,4 \}$ and continuous observations
related to EMG signals which are modeled as a multivariate Gaussian
distribution. The Myo Armband produces a $8$-dimensional spatial EMG
signal $\supscr{o}{emg}(t) \in \real^8$. We use the mean
$\supscr{\bar{o}}{emg} \in \real^8$ and standard deviation
$\supscr{\tilde{o}}{emg} \in \real^8$ of the signals over $1s$ window
and $0.2s$ frame shift as input observations.  The final feature
observed by the HMM is is given by $o :=( \supscr{\bar{o}}{emg,
  $\top$},\supscr{\tilde{o}}{emg, $\top$})^\top \in \real^{16}$.  We
collect the training data $o$ for 1 minute, during which the user
performs all $5$ gestures. The gestures are implemented in a fixed order
in a $3$ second interval for each gesture without stopping. This gives
us $12$ seconds data for each gesture spread across the $1$ minute
horizon. Next, we employ the Baum-Welch algorithm  to train the HMM model
parameters
. Details related to the Baum-Welch algorithm
implementation can be found in our previous work \cite{AS:16}. Now we
have constructed the HMM model from the training data and we can
proceed to decode the gestures in real time.
   After implementing the Baum-Welch algorithm we obtain the model
  parameters which can be used to construct the HMM. Then we use the
  standard forward algorithm to perform live decoding of the gestures
  similar to our previous work in \cite{AS:16}.
 Now, we will now look into decoding the arm
movements to complete the intention decoder. 

\subsubsection{Arm movement decoding using a Kalman filter}
We use a standard discrete-time Kalman filter~\cite{ST-WB-DF:05} to
decode the arm state $\arm(t)$ from the IMU signals
$\supscr{o}{imu}(t)$. We consider only planar motions of the arm as we
will be using a planar environment for the GUI and the formation
controller. The arm state is transformed into mouse position $m^p(t)
\in \real^2$ and velocity $m^v(t) \in \real^2$ by an appropriate
scaling and sent to the GUI (Section~\ref{sec:GUI}).  We use a
discrete, linear time-invariant model to describe the dynamics of the
mouse state, $m^p(t)$ and $m^v(t)$, based on Newton's second law. In
this way,
\begin{equation}
  \begin{bmatrix}
    m^p(t+1) \\ m^v(t+1)
  \end{bmatrix} = \begin{bmatrix}
    1 & \eta \\ 0 & 1
  \end{bmatrix} \begin{bmatrix}
    m^p(t) \\ m^v(t)
  \end{bmatrix} + \begin{bmatrix}
    \eta^2/2 \\ \eta
  \end{bmatrix} m^a(t) + w^p(t),
  \label{eqn:KF_proc}
\end{equation}
where $m^a(t)$ is the input acceleration given by the planar angular
orientation of the arm which is under our control, $\eta$ is the
update time constant and $w^p(t)$ is the Gaussian process noise.  In
this way, the acceleration of the mouse pointer is controlled by
changing the arm orientation, which is a more stable signal than the
one provided by the accelerometer.  The measurement model which uses
$\supscr{o}{imu}(t)$ signals to observe the states is given by
\begin{equation}
  y^m(t)=\supscr{r}{arm}\mathbf{I}_4 \supscr{o}{imu}(t) + w^m(t),
  \label{eqn:KF_measurement}
\end{equation}
where $\supscr{r}{arm}$ is the distance between the MYO armband to the
tip of the user's finger, which can be measured or fixed approximately
and $w^m$ is the Gaussian measurement noise present in the gyroscope
and magnetometer signals. Equations~\eqref{eqn:KF_proc}
and~\eqref{eqn:KF_measurement} are in the standard form to apply the
Kalman filter to estimate the mouse state which is then used by the
GUI program to control the mouse movement in the computer. This
enables the armband to essentially replace the computer mouse as a
complete Human Computer interaction (HCI) device, which can be used
for other purposes as well. This gives the user the opportunity to
interact with the computer using both the mouse and the
armband. Section~\ref{sec:results_myo} shows the results of our
proposed intention decoder. The decoded intentions are sent to the GUI
which is illustrated in Section~\ref{sec:GUI}.


\subsection{ Problem 5: User Interface Design}
\label{sec:GUI}

\begin{figure}
\centering
{\includegraphics[width=0.90\linewidth]{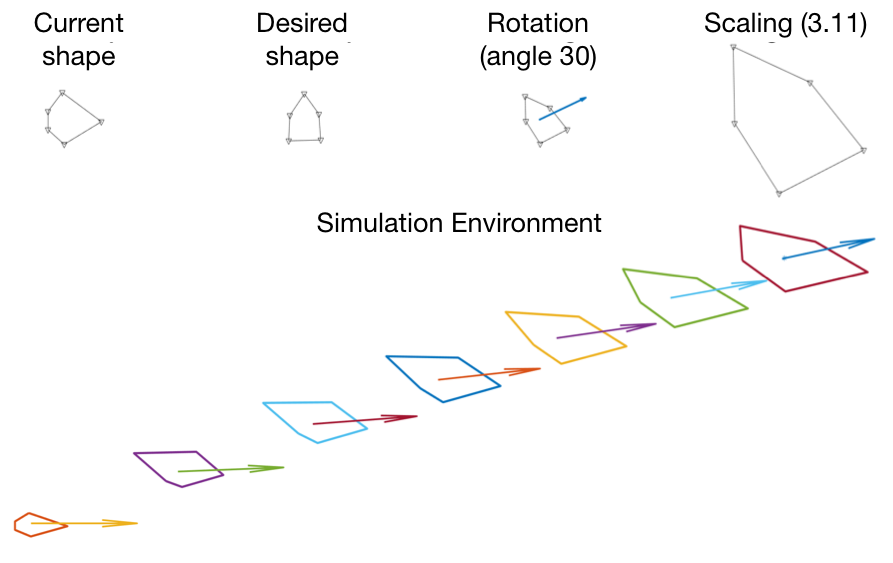}
\caption{UI used to interact with the interpreter. }
\label{fig:gui}
}
\end{figure}

We developed a GUI in MATLAB which takes in the input from the human
through the computer mouse and performs the desired behavior with
simulated robots. The user interacts with the GUI using arm movements
and gestures which are mapped to mouse movements and mouse clicks
according to Section~\ref{sec_gesture_decoding} and
Figure~\ref{fig:myo_loop}. Figure~\ref{fig:gui} illustrates a snapshot
of the GUI during the planning phase which has 5 different boxes,
whose selection will be triggered by hovering over to the desired area
with the mouse pointer. 
The current shape of the swarm is illustrated
on the top left corner of the screen. The user specifies the desired
shape $S^d$ on the $\supscr{2}{nd}$ to left side of the screen by
choosing the vertices of the polygon using arm movements and the fist
gesture or left click. Next the user proceeds to choose the rotation
$\theta^d$ on the $\supscr{2}{nd}$ to right side of the screen using
mouse scroll or the ``wave up'' and ``wave down'' gestures to increase
or decrease the angle $\theta^d$ respectively. On the top right corner
scaling $\scale$ is chosen again by the ``wave up'' and ``wave down''
gestures in a similar manner as the desired angle. The larger area in
the bottom half of the screen represents the environment where the
planning and execution of formation control takes place. The user
decides the desired centroid $c^d$ by making a ``fist'' or clicking
the left mouse button. In this manner the Human communicates their
desired intention which is sent to the interpreter that is described in
Section~\ref{sec:planner}.

\subsection{Problem 4: Swarm Controller}
\label{sec:swarm_controller}
Our swarm controller is designed to achieve the interpreter's
intention 
$v^s(l) := \{z(l),\scale(l),\cent(l),\theta(l),\mu(l)\}$ at time $l
\subscr{\tau}{int}$. Having second-order integrator dynamics for the
agents,  and the need of controlling the swarm
centroid motivates our controller which extends~\cite{JC:08-acc} (for
first-order agents) with the dynamic consensus feedback
interconnection of~\cite{MZ-SM:08a}. 
 
With $p_i,v_i$ being the position and velocity of the $\supscr{i}{th}$
agent, our second-order distributed swarm controller takes the form:
\begin{subequations} \label{eqn:sc_ind}
\begin{align}
  p_i(t+1) =&p_i(t) +v_i(t),  \label{eqn:sc_ind_p}\\
  v_i(t+1) =&-\alpha (p_i(t)+v_i(t))\ + \nonumber \\
  & \frac{\alpha}{d_i(t)}\sum_{j\neq i}\{a_{ij}(t)(p_j(t)+v_i(t))+
 \nonumber \\
\scale(l)d_{i}(t)&(z_i(l)-z_j(l))R^{\theta}(t)\} -
k^p(\agentcen{i}(t+1) - \cent(l)),\nonumber \\
  \agentcen{i}(t+1)=&\agentcen{i}(t)+ \nonumber \\
 & \sum_{j\neq
    i}w_{ij}(\agentcen{j}(t)-\agentcen{i}(t)) + p_i(t) - p_i(t-1) \label{eqn:sc_ind_c},
\end{align}
\end{subequations}
where $k^p,\alpha \in (0,1)$ are control gains and $R^{\theta}$ is the
rotation matrix corresponding to $\theta$. The variable
$\agentcen{i}(t) \in \mathbb{R}^n$ is the estimated center of the
swarm by the $\supscr{i}{th}$ agent.  Note that the $w_{ij}$ are the
Metropolis weights defined in Section~\ref{sec:graph_theory}.  This
algorithm, which applies to second-order systems, cancels out the
drift observed in \cite{JC:08-acc} with the help of dynamic consensus,
and drives the swarm to the desired centroid at time
$l\subscr{\tau}{int}$. The FODAC algorithm in~\cite{MZ-SM:08a} in
equation~\eqref{eqn:sc_ind_c} is used to distributively estimate the
mean of time varying reference signal $p(t)$ which would give us the
estimate of the swarm's centroid $\estcen(t)$.

Using \eqref{eqn:sc_ind} the swarm achieves the desired
interpreter's intention $v^s(l)$.  After some calculations, with $X(t)
= [p(t)^\top,v(t)^\top,$ $\estcen(t)^\top,q(t)^\top]^\top \in
\mathbb{R}^{4M\times n}$ as the combined state of the swarm, the state
space form of our swarm controller is represented as:
 \begin{align}
   &X(t+1)=\mathbf{A}X(t)+\mathbf{F}, \label{eqn:sc_ss} \\ \nonumber
   &\mathbf{A}=\begin{bmatrix}
     \mathbf{I} & \mathbf{I} & \mathbf{O} & \mathbf{O} \\
     -\alpha D_\mu^{-1}L_\mu -k^p\mathbf{I}& -\alpha D_\mu^{-1}L_\mu & -k^pW & k^p\mathbf{I} \\
     \mathbf{I} & \mathbf{O} & W_\mu & -\mathbf{I} \\ \mathbf{I} & \mathbf{O} & \mathbf{O} & \mathbf{O} \\
   \end{bmatrix}, \\ \nonumber &\mathbf{F}= [\mathbf{O}^\top ,\ [\scale
   \alpha D_{\mu}^{-1}L_{\mu}zR^{\theta} +k\ones[M] \cent]^\top \ ,\
   \mathbf{O}^\top \ ,\ \mathbf{O}^\top]^\top.
\end{align}   
Here $q(t)=p(t-1)$ is a dummy state introduced to obtain a linear
system in standard form. It is interesting to note that the swarm
controller \eqref{eqn:sc_ss} consists of an autonomous component
$\mathbf{A}$ and a controlled component $\mathbf{F}$ housing the
desired interpreter's intention $v^s(l)$. So $v^s(l)$ can be
communicated once at the beginning of the $\supscr{l}{th}$ iteration
and the agents just need to adjust their positions and communicate
locally with their neighbors to achieve the intermediate goal.
Letting
$Z^d(l)=[\ones[M]\cent(l)]^\top+[\scale(l)z(l)R^\theta(l)]^\top$, the
desired intention $X^d(l) \in \mathbb{R}^{4M\times n}$ in this state
space is given by
$X^d(l)=[Z^d(l)^\top,\mathbf{0}^\top,[\ones[M]\cent(l)]^\top,Z^d(l)^\top]^\top$.
Now we will theoretically analyze the performance of the proposed
swarm controller in the next section.

\subsection{Swarm Controller Analysis}
\label{sec:swarm_analysis}  
In this section we will analyze our proposed controller
\eqref{eqn:sc_ss} to determine stability and convergence. We will look at the case when $\mathcal{G}_\mu(t)$ remains constant for
$t \in [\tau^l(l-1),\tau^l(l)]$.
   So this makes our system
time-invariant in that interval.  In this work, we will make use of
the following assumptions on $\mathcal{G}_\mu(t)$:
\begin{assumption}[Connectivity]
  The communication graph $\mathcal{G}_\mu(t)$ has at least one
  globally reachable vertex at every time $t$.
\end{assumption}
\begin{assumption}\longthmtitle{Constant graphs}
  The communication graph $\mathcal{G}_\mu(t)$ remains constant for $t
  \in [(l-1) \subscr{\tau}{int},l\subscr{\tau}{int}]$.
\end{assumption}

System~\eqref{eqn:sc_ss} represents $n$ copies of the same dynamics
corresponding to $n$ different dimensions. To simplify notation, we
will analyze only one of the dimensions. After fixing $\mu$ and
omitting it for simplicity, our swarm controller~\eqref{eqn:sc_ss} can
be reduced by combining the $p$ and $v$ dynamics to obtain:
\begin{subequations} \label{eqn:sc_proof}
  \begin{align}
    p(t+1)=& (\mathbf{I}-\alpha D^{-1}L)p(t) - k^p\ones[M]\estcen(t) + \mathbf{F}_1(l), \label{eqn:sc_proof1}\\
    \estcen(t+1)=& W\estcen(t)+ p(t) - q(t), \label{eqn:sc_proof2}\\
    q(t+1)=& \label{eqn:sc_proof3} p(t).
  \end{align}
\end{subequations}
where $\mathbf{F}_1(l)=\scale(l)\alpha D^{-1}Lz(l)R(l) +
k^p\ones[M]\cent(l)$. System~\eqref{eqn:sc_proof} is an interconnected
system whose stability depends on the chosen gains $\alpha$ and
$k^p$. We will use the discrete analogue of composite Lyapunov
functions~\cite{HKK:02}  to design the gains that guarantee the
stability of the interconnected system. With $\delta_1=1-(1-\alpha
\lambda_2^N)^2 ,\ \ \delta_2=1-(1-\lambda_2^W)^2$ we can state the
following theorem.

\begin{theorem}\longthmtitle{Stability of Swarm Controller}
\label{thm:stab}
Under Assumption~1 (connectivity) and Assumption~2 (constant
interconnection graph), with the control gains satisfying $k^p <
\frac{\delta_1 \delta_2}{2}$, the swarm globally uniformly
asymptotically stabilizes to the desired state $X_d$ under the swarm
controller dynamics \eqref{eqn:sc_ss} from any initial condition.
\end{theorem}

The proof of Theorem~\ref{thm:stab} is presented in the Appendix. Next
we will use the results of Theorem~\ref{thm:stab} to get an intuition
of the role of graph connectivity ($\lambda_2^N$ and $\lambda_2^W$) in
the convergence of our swarm controller \eqref{eqn:sc_ss}.

\begin{corollary}
\label{corr:1}
The convergence rate of~\eqref{eqn:sc_ss} is directly proportional to
$\lambda_2^N$ and $\lambda_2^W$ of the communication graph.
\end{corollary}

The proof of Corollary~\ref{corr:1} can be found in the Appendix. Using these results we will design a planning algorithm, which optimally determines the intermediate subgoals which will be described in Section \ref{sec:planner}.

\subsection{The interpreter}

In this section we describe the role of the interpreter in the
framework. For ease of illustration, we consider the formulation in
$2$D space. The interpreter mainly consists of two parts: the
behavior specifier and the planner, which are illustrated in the
following paragraphs.
\subsubsection{Problem 2: Behavior Specifier}
The Behavior specifier converts the desired human intention into
parameters that can be comprehended by the swarm. The human user
specifies the desired shape $S^d \in \mathbb{P}$ which takes the form of an arbitrary polygon, the
desired centroid $\cent^d \in \real^2$, scaling $\scale^d \in \real$ and
rotation $\theta^d \in \real$. The interpreter then decides the
formation denoted by the relative positions of the agent $z^d\in
\real^{M\times n}$, which would best illustrate the shape $S^d$ given
by the human. For simplicity, we use a uniform distribution in the
interior of the shape $S^d$ to obtain $z^d$, which is illustrated in
Figure~\ref{fig:formation_spec}.
The human specifies the polygon by
providing the vertices sequentially using the GUI from
Section~\ref{sec:GUI}, which is shown on the left side of the
Figure~\ref{fig:formation_spec}. The corresponding formation density
$\rho^M=M/ \area(S^d)$ is calculated, where $\area(S^d)$ is the area
of polygon $S^d$. We assume the density is large enough to fit $M$
robots in the shape $S^d$. Note that, since the shape $S^d$ is
bounded, there exists a large enough box $B$ such that $S^d
\subseteq B$ and $M \frac{\area(B)}{\area(S)}$ is equal to a perfect square
$r^2$, for some $r^2 \ge M$ and $r^2 /\area(B) = M / \area(S) =
\rho^M$.  
  Using this density, robots are distributed uniformly in the
bounding box $B$ of the polygon $S^d$ by creating a meshgrid. Finally,
we discard the generated points not in the polygon and we arrive at
the desired formation $z^d$ of $M$ points shown in the right half of
Figure~\ref{fig:formation_spec}.
 The parameters $z^d$, $S^d$, $\cent^d$, $\scale^d$ and $\theta^d$ are
 passed on to the Planner, which is described next.


\subsubsection{Problem 3: Planner}
\label{sec:planner}
The Planner receives the decoded human intention in the form of
desired formation ${S^d}$ (or, equivalently, $z^d$), scaling
$\scale^d$, rotation $\theta^d$, and centroid $\cent^d$.  The planner
then constructs a set of intermediate way points
$\{S(l),\scale(l),\theta(l),\cent(l)\},\; \forall l \in \until{N}$,
where $N$ denotes the number of intermediate steps in the plan to
reach the final goal. 
  
To do this, we employ an $N$-Horizon Discrete Switched Linear
Quadratic Regulator (DSLQR) formulation. A particular DSLQR problem
with a dynamical variable $h \in \real^d$ and time horizon $l \in
\until{N}$ can be formulated as follows: 
\begin{subequations}
\label{eqn:DSLQR_prob}
\begin{align}
  \min J(u,\mu)=& \sum_{l=0}^N (h(l)^\top Q_\mu h(l) + u(l)^\top R_\mu u(l)) \nonumber \\
  &+ h(N)^\top Q_f h(N), \label{eqn:DSLQR_cost} \\
  \text{subject to}\ & h(l+1)=\mathcal{A} h(l) + \mathcal{B}
  u(l), \label{eqn:DSLQR_dyn}
\end{align} 
\end{subequations}
where $h(0) = h_0$.  Here, the
running cost consist of a switching LQ cost function, with
parameterized matrices $Q_\mu$ and $R_\mu$, depending on a mode $\mu$.
The function will be designed to enhance swarm performance while the
linear constraint will be used to enforce an easy-to-interpret
behavior by a human, which defines a Human Interpretable (HID)
dynamics.

Details and methodology of DSLQR systems can be found
in~\cite{WZ-JH-AA:09}. We show next how we apply this approach in our
particular setup and describe the matrices that we choose for our
framework.
 
(i) \textit{Human-Interpretable Dynamics:} We introduce the notion of
Human Interpretable Dynamics (HID) to denote a dynamical system that
can be easily understood by a human.  Since the interpreter needs to
provide feedback to the user, the planner needs to provide an
abstraction of the complicated swarm dynamics in an $Mn$-dimensional
space. 
These dynamics need to be slower than the swarm dynamics to enhance
human interpretability, and are hence implemented in the $l$ timescale
described in Section~\ref{sec:framework}.
  
Here, we propose a simple linear dynamical system approach to model
these dynamics, which takes into account the desired human intention
$h^d = (S^d,\scale^d, \theta^d, \cent) $. We suppose that fully
actuated linear dynamical systems are more easily understandable by
humans, as opposed to other nonlinear system models.  We let
$h=[S,\scale,\theta,\cent]^\top$ denote the state of the HID system
with $h(l) \in \mathbb{H}$, where $\mathbb{H}=\mathbb{P}\times
\real^n \times \real \times
\real$. 
Then, the HID takes the form:
\begin{equation}
  h(l+1)=\mathcal{A}h(l)+\mathcal{B}u(l),
  \label{eqn:HID}
\end{equation}
where matrices $\mathcal{A},\mathcal{B} \in \mathbb{H}\times
\mathbb{H}$ and control input $u\in \mathbb{H}$. In this paper, we
choose $\mathcal{A}$ and $\mathcal{B}$ to be identity matrices. This
seems to be the most intuitive dynamics as the control input applies
directly on the system. In future work, we will study alternative
choices for these dynamics.

\begin{figure} 
\centering
\subfigure[\scriptsize{HID illustration}]
{\includegraphics[width=0.24\textwidth]{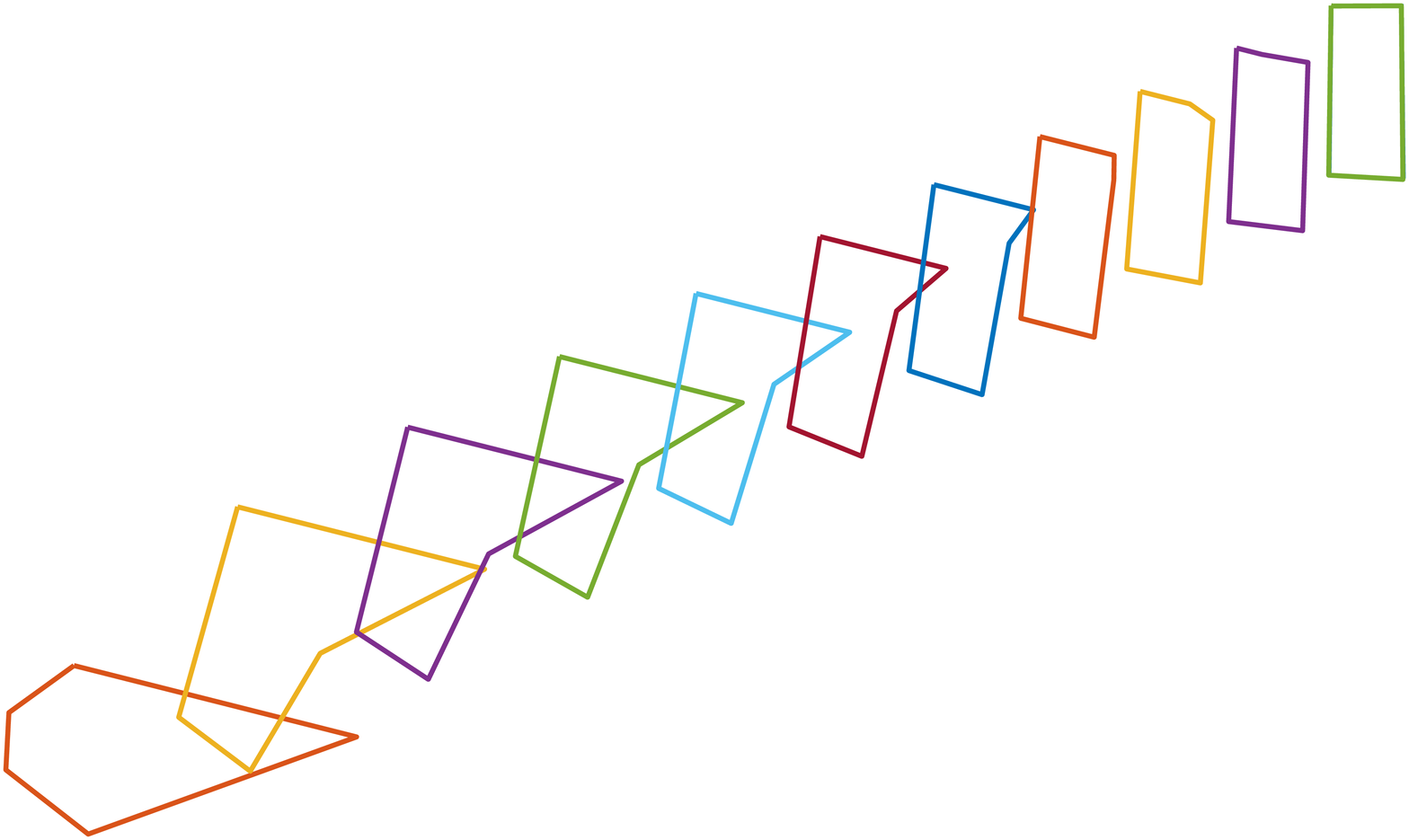}\label{fig:HID_spec}} 
\subfigure[\scriptsize{Formation Specifier}]
{\includegraphics[width=0.24\textwidth]{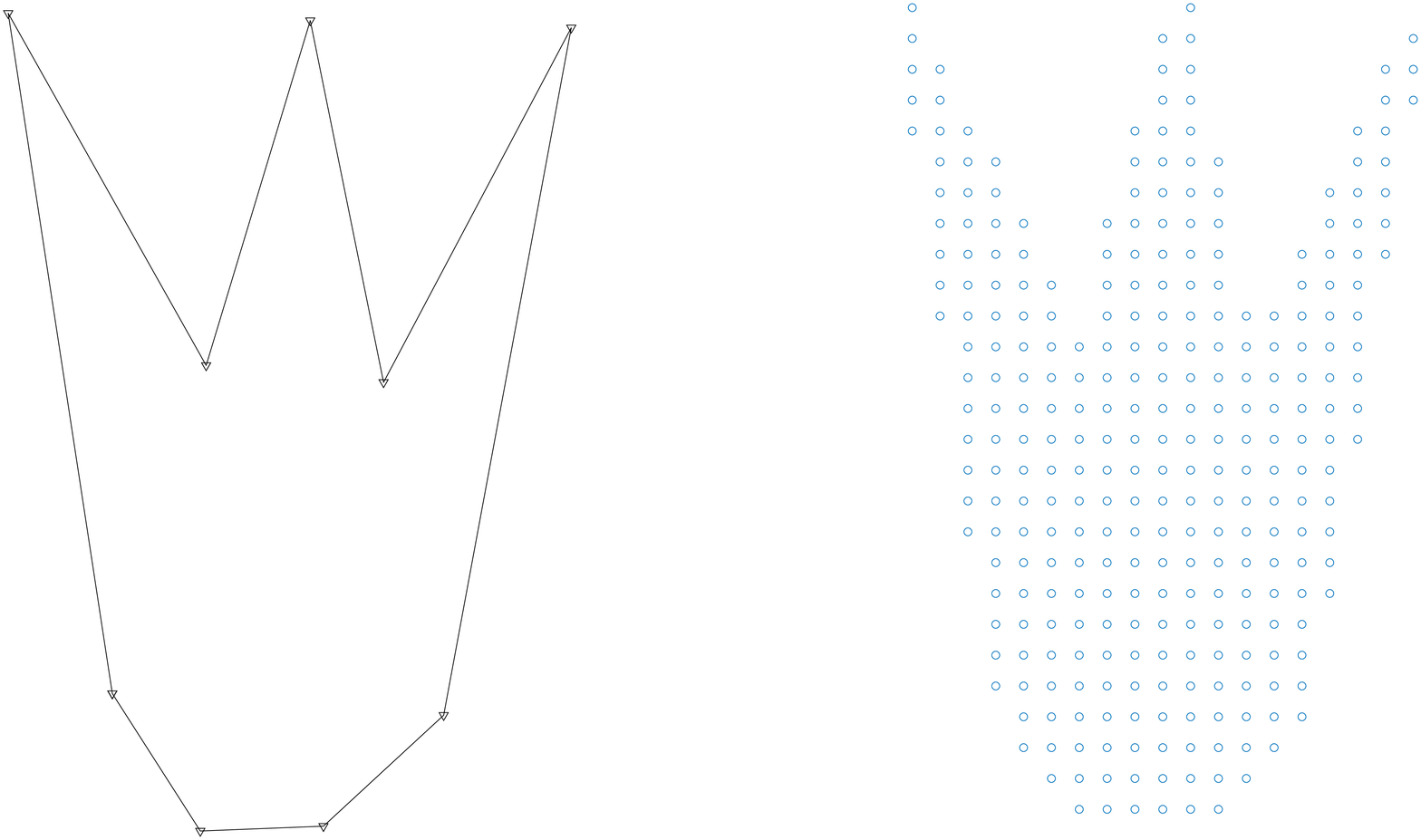}\label{fig:formation_spec}}
  \caption{(a) HID illustration for shape changing from rotated cone to a
    standing rectangle. The model parametrs used are
    $\mathcal{A}=\mathcal{B}=Q=\mathbf{I}_h$, $R=100 \mathbf{I}_h $
    and $Q_f=1500 \mathbf{I}_h$.
    (b) Left: The user specifies the
    desired shape $S^d$ by providing $v$ vertices (triangles). Right:
    the interpreter determines the relative positions $z^d$ of $M=500$
    agents (blue dots) to represent the shape drawn by user.}
\end{figure}

We use the $N$ horizon Discrete LQR control technique to drive the HID
towards $h^d$ starting from some initial configuration $h(0)=h_0$. By
considering a change of variable $h^e(l)=h(l)-h^d$, we define a first
term contributing to the problem cost functional as follows:
\begin{align}
  \subscr{J}{HID}(u)= & \sum_{l=0}^{N-1}(h^e{^\top}(l)Qh^e(l) +
  u(l)^\top Ru(l)) + \\ \nonumber & h^e(N)^{\top}Q_f h^e(N) .
\label{eqn:HIDcostfunctional}
\end{align}
where the matrices $Q,R,Q_f \in \mathbb{H}\times \mathbb{H}$ are
positive definite and $u(l),\; \forall l \in \until{N}$ is a step
change applied during the $\supscr{l}{th}$ time. So $u(l)$ is chosen
such that the cost $\subscr{J}{HID}$ is minimized. This is solved
using the standard LQR approach, and the results are shown in
Figure~\ref{fig:HID_spec} for a $N=10$ horizon problem. Intuitively,
one can choose these matrices to satisfy $Q \prec R \prec Q_f $ to
provide a more human ``interpretable'' dynamics. This condition
implies that the priority is to reach the desired behavior $h^d$ with
small changes in the intermediate steps, which would make it look more
natural and ``interpretable'' to the human eye as seen in
Figure~\ref{fig:HID_spec}. 
Figure~\ref{fig:HID_spec} shows the stages of transformation of a 5
sided polygon to a rotated and translated 4 sided polygon. The figure
depicts a seemingly natural transition which can be easily interpreted
by the user, thus justifying the HID formulation. The case of mismatch
in the number of vertices in the initial and desired shapes is handled
by adding vertices appropriately on the perimeter of the shape that
has fewer vertices.

(ii) \textit{Swarm performance costs.} We just discussed how to
generate intermediate shapes taking into account the HID. Now we
consider the swarm performance and communication cost to choose the
operating mode $\nu$ in the general
setup. 
The operating modes $\nu$ correspond to a subset of $\nu$-disk graphs
defined over the swarm when distributed over a shape.  Since agent
formations are chosen in a consistent manner as described in
e.g.~Figure~\ref{fig:formation_spec}, 
   the number of possible graphs over the agents for
different $\nu$ is very much reduced and remains constant for scaled
shapes. From now on, we consider this set is given by
$\{\nu_1,\dots,\nu_m\}$ by choosing appropriate communication radii.

\textit{Operating costs involved:}  
To increase  the speed of convergence 
and to facilitate quicker interpretation by a human, 
we need to maximize the notion of connectivity involving the second smallest eigenvalue $\lambda_2^N$ or $\lambda_2^W$ of the respective Laplacian matrices $L^N$ and $L^W$. 
This can be found from the determinant of the matrix $G \in
\mathbb{R}^{(M-1)\times (M-1)}$ defined as $G=F^\top L^NF$ with $F\in
\mathbb{R}^{M\times (M-1)}$, $F\ones[M]=0$ and $F^\top F=\mathbf{I}$ . Since the determinant of a matrix is a product of its eigenvalues, connectivity determined by
  $\lambda_2^N$ increases iff the determinant of $G$ increases. 
  So the connectivity cost $\subscr{J}{CON}(l)$
  being
in formation $z$ and operation mode $\nu$ at time $l$ is given by:
\begin{equation}
  \subscr{J}{CON}(\nu,h) = -\kappa_1\log\det(\kappa_2 G_{\nu}(l)).
  \label{eqn:Connectivity cost}
\end{equation}
  To ensure $\subscr{J}{CON}$ remains well scaled and positive we introduce positive constants $\kappa_1$ and $\kappa_2$ respectively.
  Having a $\nu$ corresponding to a higher
communication radius implies that we will be using more energy to
communicate and maintain communication links. This is encoded as a
communication cost $J_{com}(l)$ being in formation $z$ and operation
mode $\nu$ at time $l$. It is given by
\begin{equation}
\subscr{J}{COM}(\nu,h)=\kappa_3 \log(\nu_\nu^2\ones[M]^\top A_{\nu}(h)\ones[M]),
\label{eqn:Connectivity cost}
\end{equation}
where $\nu(l)$ is the communication range at time $l$ and $\kappa_3$ is a positive constant used for scaling. 

Adding these costs together defines the total cost used by the planner
as:
\begin{align}
  J(u,\nu)=&\subscr{J}{HID}(u) + \subscr{J}{CON}(\nu,h) + \subscr{J}{COM}(\nu,h) \label{eqn:modifiedcostfunctional} \\
  J(u,\nu)= & \sum_{l=0}^{N-1}(\bar{h}^e(l){^\top}
  \overline{Q}_{\nu(l)}\bar{h}^e(l) + \overline{u}(l)^\top \overline{R} \overline{u}(l) + \nonumber \\
  \nonumber & \bar{h}^e(N)^{\top}Q_f\bar{h}^e(N) .
\end{align}
where $\overline{Q}_{\nu}=\begin{bmatrix}
Q & 0 \\ 0 & \subscr{J}{CON}(\nu)+\subscr{J}{COM}(\nu)
\end{bmatrix}$ , $\bar{h}^e=\begin{bmatrix}
h^e \\ 1
\end{bmatrix}$, $\overline{u}=\begin{bmatrix}
u \\ 0
\end{bmatrix}$ and $\overline{R}=\begin{bmatrix}
R & 0 \\0 & 1
\end{bmatrix}$ 
. 
  Observe that a solution to the above problem
requires the evaluation of all possible graph combinations for
different chosen controls $u$.  By choosing the graphs based on the
communication radii, and considering a class of formations, we reduce
significantly the number of possible graphs to evaluate. In addition,
we employ the DSLQR formulation from \cite{WZ-JH-AA:09} to obtain the
optimal set of $u(l)$ and $\nu(l)$ which minimizes $J$.
 Our optimization is done in the
following sequential manner: first we optimize in the sequence of $\bar{h}^e$
and $\overline{u}$, then, given this, we optimize in the $\nu$ variable using the DSLQR approach from \cite{WZ-JH-AA:09}. This is
further illustrated and discussed in
Section~\ref{sec:results-switchedsys}.

\section{Implementation Results}
\label{sec:results}
\subsection{System Setup}
The user has the choice to use either the MYO armband or the mouse to
interact with a GUI to control the formation of a simulated swarm in a
two dimensional environment. The swarm controller developed in
Section~\ref{sec:swarm_controller} essentially generates waypoints for
the swarm to follow, we assume holonomic dynamics for the individual
agents and assume they reach their respective waypoints. We do not
  focus on collision avoidance, which will we addressed in
future work. We utilize the ROS kinetic framework with Python
scripting language to interface with the MYO armband and control the
mouse pointer. We use Matlab to create the GUI shown in
Figure~\ref{fig:gui}, which uses the mouse or the MYO armband as an
input device. For the formation controller we set the control gain $\alpha=0.15$ and proportional constant $k^p=0.03$. 
\subsection{Intention Decoding}
\label{sec:results_myo}
\begin{figure}
\centering
\subfigure[\scriptsize{Mouse movement with wearable}]
{\includegraphics[width=0.23\textwidth]{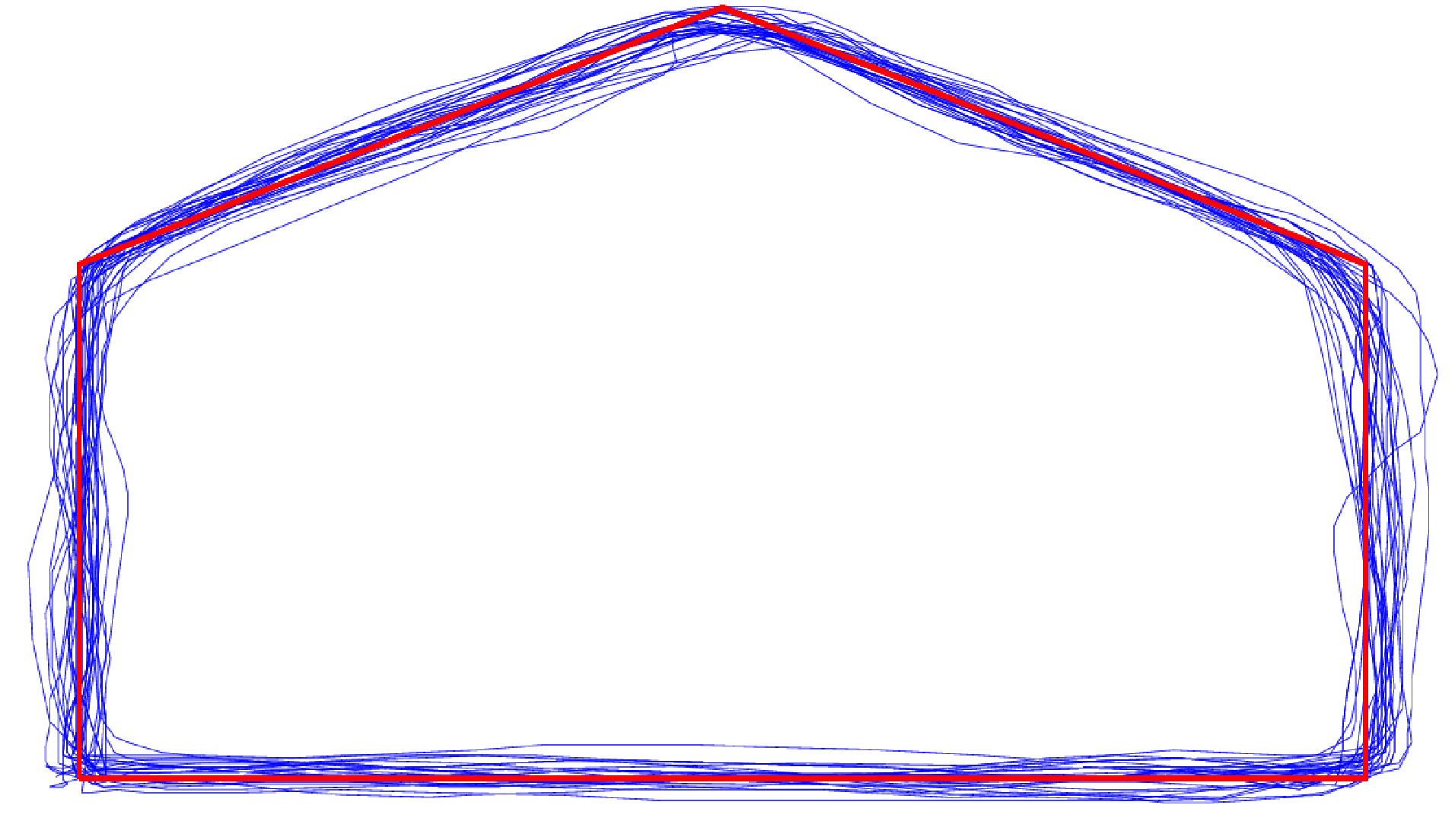}}
\subfigure[\scriptsize{Mouse movement without wearable}]
{\includegraphics[width=0.23\textwidth]{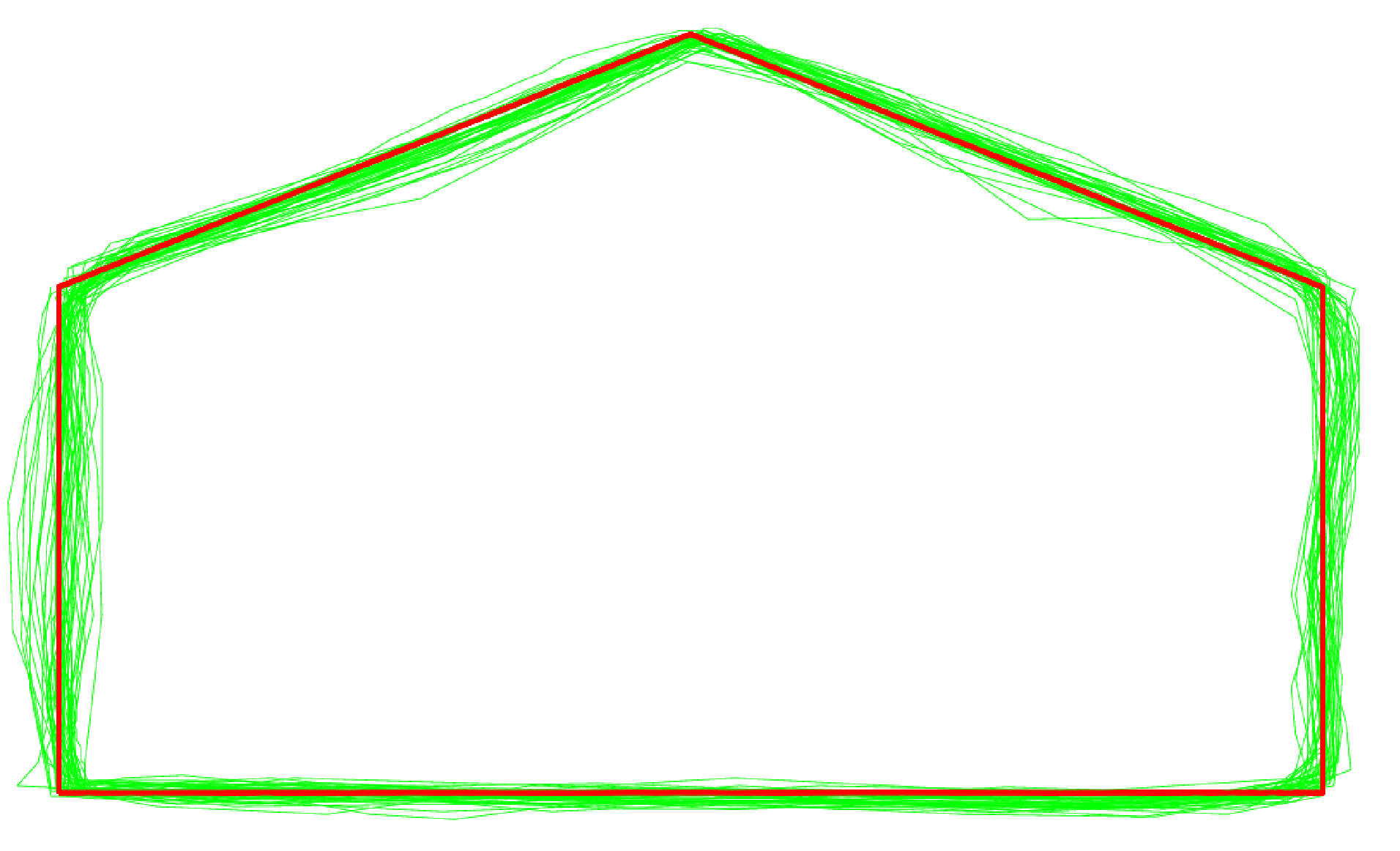}}
\caption{Aggregate results of tracing a pentagon.(Red) a) The user specifies the shape by using MYO armband. (Blue) b) The user specifies the shape by using the mouse. (Green)}
\label{fig:movement_accuracy_test}
\end{figure}

We performed tests to gauge the accuracy and speed of the proposed HMM
and Kalman Filter models. For the HMM model, some of our previous
tests had given an accuracy levels of over 90\% on an
average~\cite{AS:16} for similar gestures and framework. On preliminary tests we observed similar results and hence, in the
interest of space, we skip this accuracy test for the HMM model. 
For the effectiveness of the arm movement decoder, we compare the
results of operating a mouse with and without the MYO
armband. Figure~\ref{fig:movement_accuracy_test} represents the
aggregate results over 5 trials. The user was tasked to continuously
trace a pentagon which represents the human intention for a minute. It
can be seen from the Figure~\ref{fig:movement_accuracy_test} that the
results are similar for both cases. Table
\ref{tab:intention_decoder_comparison} describes the error involved in
each of the trials. It can be seen that the errors involved are about
the same with both interfaces, however the speed of using the mouse is
higher than the other. This is also due to the fact that users are
accustomed to using the mouse for years and need time to adapt to the
new interface. But in the $\supscr{5}{th}$ trial it can be seen that
the performance with the wearable matches many trials with the mouse,
which shows that the user can adapt quickly to use the new interface.

\begin{table}[]
\centering
\caption{Error comparison mouse and wearable.}
\resizebox{0.49\textwidth}{!}{%
\begin{tabular}{ccccccc}
       & \multicolumn{3}{c}{\textbf{Mouse}} & \multicolumn{3}{c}{\textbf{Wearable}} \\
Sl. no & Loops  & Avg Error  & Total Error  & Loops   & Avg Error   & Total Error   \\
1      & 7      & 0.026      & 122.57       & 5       & 0.038       & 179.26        \\
2      & 8      & 0.028      & 129.40       & 5       & 0.037       & 174.74        \\
3      & 9      & 0.031      & 147.02       & 7       & 0.048       & 222.07        \\
4      & 8      & 0.031      & 148.92       & 7       & 0.05        & 235.27        \\
5      & 9      & 0.035      & 161.50       & 5       & 0.029       & 132.72       
\end{tabular}%
}

\label{tab:intention_decoder_comparison}
\end{table}

\subsection{DSLQR Formulation} 
\label{sec:results-switchedsys}
Now we will validate the proposed framework by running simulations of a swarm of 50 agents to
reach the desired human intention. Below, we illustrate a particular execution of our framework.

%

\begin{figure}
\centering
\subfigure[\scriptsize{Current shape}]
{\includegraphics[width=0.06\textwidth]{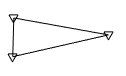}}
\subfigure[\scriptsize{Desired Shape}]
{\includegraphics[width=0.06\textwidth]{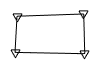}}
\subfigure[\scriptsize{Desired rotation: $\theta^d=~50^{\circ}$}]
{\includegraphics[width=0.15\textwidth]{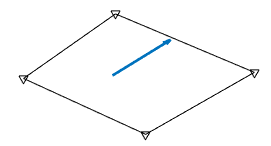}}
\subfigure[\scriptsize{Desired scaling: $s^d=~11.6$}]
{\includegraphics[width=0.15\textwidth]{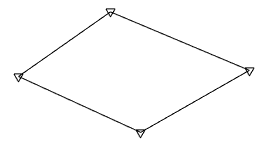}}
\subfigure[\scriptsize{Planning and Execution}]
{\includegraphics[width=0.40\textwidth]{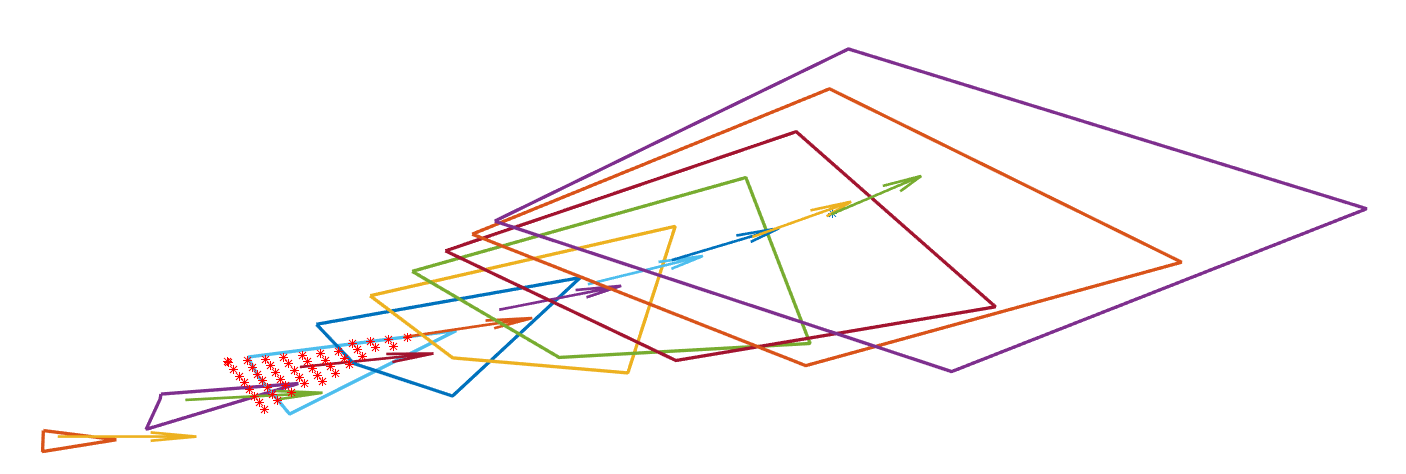}\label{fig:results_plan}}
\subfigure[\scriptsize{Switching cost throughout execution}]
{\includegraphics[width=0.40\textwidth]{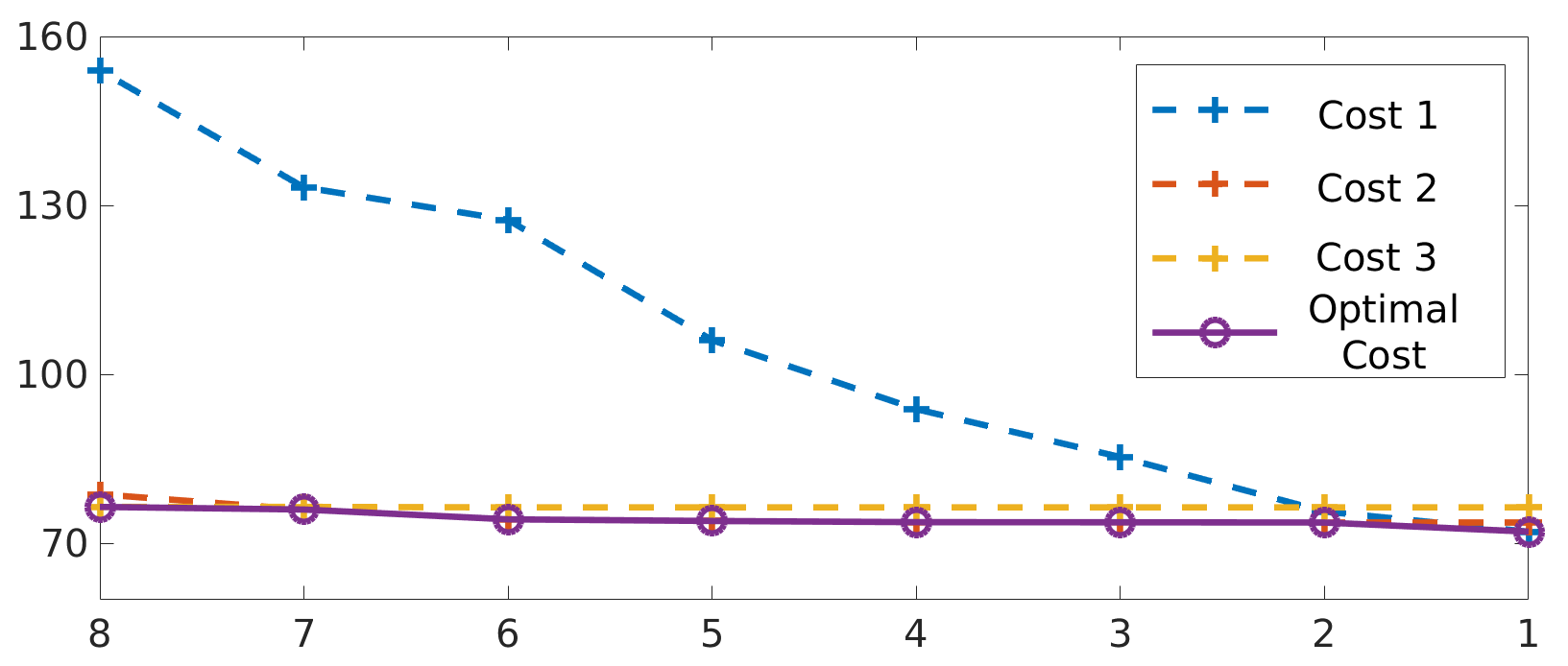}\label{fig:switchingcost}}
\subfigure[\scriptsize{Formation error $e^f(t)$ }]
{\includegraphics[width=0.24\textwidth]{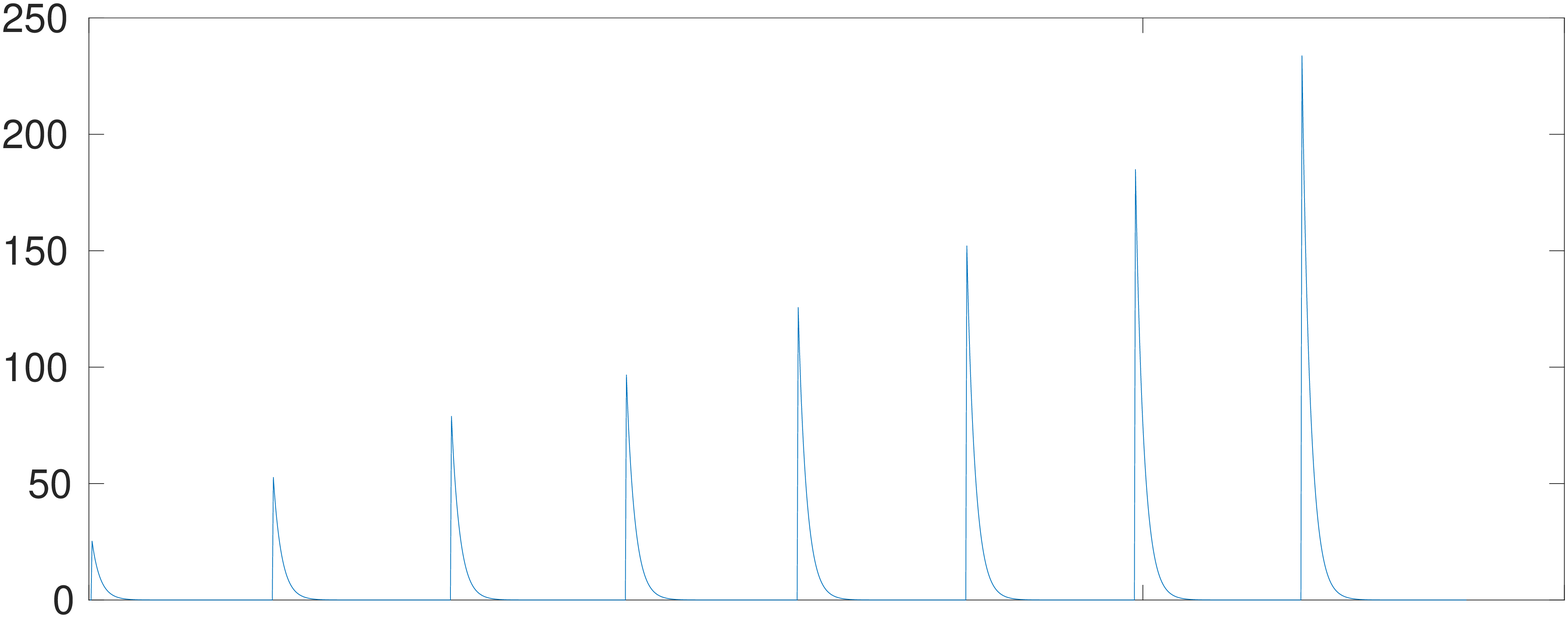}\label{fig:results_shape}}
\subfigure[\scriptsize{Centroid error $e^c(t)$ }]
{\includegraphics[width=0.24\textwidth]{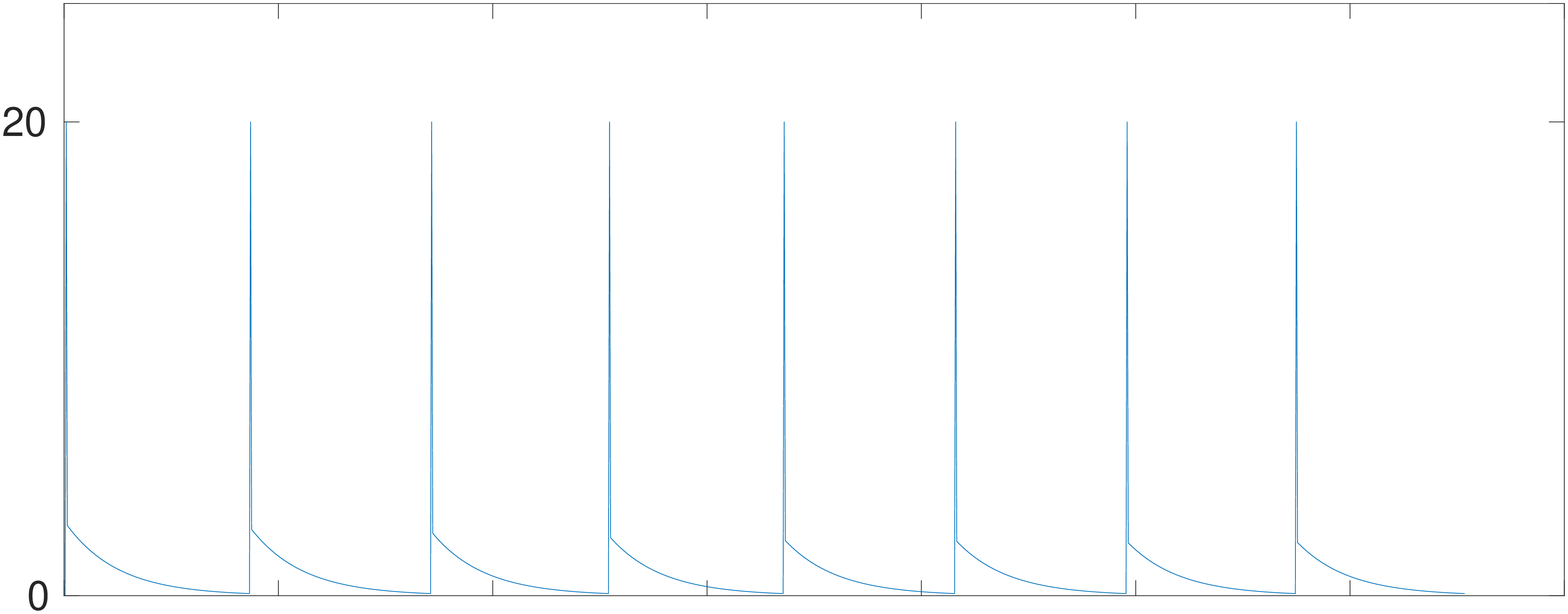}\label{fig:results_cent}}
\caption{Results of executing a particular desired behavior communicated by the human.}
\label{fig:sim_results}
\end{figure}

 Figure \ref{fig:sim_results}(a)-(d) indicate the desired human intention communicated by the human. Using $\mathcal{A}=\mathcal{B}=Q=\mathbf{I}_h$, $R=100 \mathbf{I}_h $, $Q_f=1500 \mathbf{I}_h$, $\kappa_1=10^6$, $\kappa_2=0.05$, $\kappa_3=2\times 10^4$ the planner was implemented for a
$N=8$ horizon problem with $m=3$ subsystems. The communication ranges are $\nu(l) \in \{10,40,150\}$, corresponding to the three operating modes. Figure~\ref{fig:results_plan} illustrates the intermediate shapes resulting from the $8$ horizon planner, starting from the current intention(triangle on the left), to the desired intention(larger rotated quadrilateral) on the right. The intermediate shapes look natural and the progression is gradual and intuitive, which justifies the notion of HID.
Figure~\ref{fig:switchingcost} describes the evolution of the cost \eqref{eqn:modifiedcostfunctional} and switching strategy in a backward horizon. We can see that switching occurs in a timely manner to maintain minimum costs according to \eqref{eqn:modifiedcostfunctional}. Switching occurs from $\supscr{1}{st}$ mode to the
$\supscr{2}{nd}$ mode during the $\supscr{2}{nd}$ timestep. During the $\supscr{7}{th}$ timestep another switching occurs to the $\supscr{3}{rd}$ operating mode to maintain minimum cost. This is coherent with the intuition of using larger communication radii for more sparse swarms. As the scaling increases with every timestep the agents are forced further apart and the cost of using a smaller communication range $\nu=10$ rapidly increases. Whereas, the cost of using the largest range $\nu=150$ remains almost constant throughout because the connectivity and communication costs mostly remain the same.
Figure~\ref{fig:results_plan} shows the execution of the swarm controller during the $l=2$ horizon. Each of the red dots represent individual agents of the swarm. We evaluate the performance of the swarm controller~\eqref{eqn:sc_ss} by measuring the error with respect to the intermediate formations and centroid at each time step $t$. The formation error and centroid error are measured as $e^f_l(t)=\|p(t)-s(l)z(l)R(\theta(l))\|$ and $e^c_l(t)=\|c(t)-c^d(l) \|$ respectively in reaching the $\supscr{l}{th}$ intermediate goal. The evolution of these errors(y-axis) with respect to time $t$(x-axis) is illustrated in Figures~\ref{fig:results_shape} ~and~\ref{fig:results_cent}. We see that the swarm successfully reaches every intermediate goal and finally reaches the desired human intention.
  


%

\section{Conclusions and Future Work}
In this work we have proposed and successfully implemented a novel HSI
framework for formation control, where the user draws the desired
shape using intuitive gestures, and the swarm successfully depicts the
drawn shape. We have combined diverse tools from control theory,
network science, machine learning, signal processing, optimization and
robotics to create this multi-disciplinary framework. Firstly, we have
demonstrated the effectiveness and intuitiveness of human interaction
using this framework, whose accuracy and speeds are comparable to
standard interaction devices. Next, we have proposed and utilized a
unique notion of human interpretable dynamics along with switching
systems to plan intermediate natural shapes for the swarm to depict,
which can be easily understood by the human and the swarm. We have
also developed, analyzed and illustrated a novel decentralized
formation controller capable of reaching any shape and centroid in the
$2-D$ space. Lastly, we have integrated the framework by developing a
GUI environment which interacts with user by means of gestures, and
rest of the framework is encapsulated in the GUI using matlab
simulations.
 
Future work will involve validation of the proposed framework with robustness towards noise and uncertainties. We also wish to learn the Human Interpret-able dynamics from existing human behavior models and data. 

\label{sec:conclusion}

\begin{ack}
  We thank
  Mac Schwager for useful discussions regarding the HMM formulation
  used in this work. We also thank Chidi Ewenike, Ramon Duran and Tomaz Torres for their help
  in developing the Myo armband setup used in this work.
\end{ack}

\section*{Appendix}

\subsection*{Preliminaries for proof of Theorm \ref{thm:stab}}
Let us first define the following quantities :
$x=(x_1^\top, x_2^\top, x_3^\top)^\top$, $ x_1=p,\ x_2=\estcen,\ x_3=q,\
\mathbf{F}_2=\mathbf{F}_3=\mathbf{0}, \in \mathbb{R}^M$ ,
$f_1(x_1)=(I-\alpha D^{-1}L)x_1;\ \ f_2(x_2)=Wx_2;\ \ f_3(x_3)=\mathbf{0}$
$g_1(x)=-k^px_2; \ \ g_2(x)=x_1-x_3; \ \ g_3(x)=x_1$ ; and
$\mathbb{F}=[\mathbf{F}_1^\top,\mathbf{F}_2^\top,\mathbf{F}_3^\top]^\top$.

With these definitions System~\eqref{eqn:sc_proof} can be
represented as :
\begin{equation}
  x_k(t+1)=f_k(x_k(t)) + g_k(x(t)) + \mathbf{F}_k, \ \ \forall k \in \{1,2,3\},
\label{eqn:sc_proof_comp-lyap}
\end{equation}
where $f_k$ is the system dynamics of the $\supscr{k}{th}$ system,
$g_k$ is the interconnection to the $\supscr{k}{th}$ system and
$\mathbf{F}_k$ is the drift of the $\supscr{k}{th}$ system. Now
$f_1(x_1)+\mathbf{F}_1$ resembles the shape stabilizing JOR algorithm
in \cite{JC:08-acc} with some additional centroid drift
$k^p\ones[M]\cent$. From \cite{JC:08-acc} we know this system
converges to the desired shape with some centroid translation.
Henceforth, we will ignore the drift $\mathbb{F}$ while analyzing the
overall system stability. As we see next, stability is established by
first analyzing the convergence rates of each of the subsystems
defined by $f_k$, and by identifying suitable conditions on the
interconnections $g_k$, for $k \in \{1,2,3\}$.  To this end, we define
the Lyapunov function $V(x_k) =\frac{1}{2} x^\top_k x_k $, defined
over $x_k$ for $k \in \{
1,2,3\}$.

\begin{lemma}
  The subsystem $x_1(t+1) = f_1(x_1(t)) - \ones[M]
  \ones[M]^\top]x_1(t)$ is globally
  uniformly asymptotically stable at $x_1=\mathbf{0}$. 
\label{lem:f_1_stab}
\end{lemma}
\begin{pf}
  Considering $A=I_M-\alpha D^{-1}L$, we have that the eigenvalues
  $\lambda^A \in (0,1]$ and $1$ is a simple eigenvalue with right
  eigenvector $\ones[M]$, which shows $x_1(t+1)=Ax_1(t)$ is globally
  stable.  We can perform a similarity transformation on $A$ to get
  $A_s=I_M- \alpha L^N$ where $L^N=D^{\frac{-1}{2}}LD^{\frac{-1}{2}}$
  is the symmetric normalized Laplacian of the graph.  It holds that
  the eigenvalues of $A_s$ are the same as $A$ and the eigenvectors
  are those of $A$ scaled by a factor of $D^{\frac{-1}{2}}$. We
  perform a Hotelling deflation~\cite{YS:03} on $A_s$ using the
  largest eigenvalue to get $\overline{A}=A_s-D^{ \frac{-1}{2}}
  \ones[M] \ones[M]^\top D^{ \frac{-1}{2}}$.  In this way, we have
  deactivated the largest eigenvalue of $A_s$ and now we have
$\lambda^{\overline{A}} \in [0,1- \alpha \lambda_2^N]$ 
    where $\lambda_2^N$ is
  the second smallest eigenvalue of the normalized Laplacian
  $L^N$. We will proceed by analyzing the
  stability properties of $\overline{A}$ which is similar to analyzing
  the stability of $x_1(t+1)=[(I_M-\alpha D^{-1}L)- \ones[M]
  \ones[M]^\top]x_1(t) $ since the eigenvalues and their related
  properties are the same. 

  With $\Delta V(x_1)= V(x_1(t+1)) - V(x_1(t))$ and
     $Q=\overline{A}^\top \overline{A} - I_M$ we have
  \begin{equation*}
    \Delta V(x_1) = x_1^\top Q x_1 < 0  .
  \end{equation*}
The above observation follows from the fact that $\overline{A}$ is
symmetric and $\lambda^{\overline{A}} \in (0,1- \alpha \lambda_2^N]$,
hence the eigenvalues $\lambda^Q \in (-1,(1-\alpha \lambda_2^N)^2
-1]$, which makes $Q$ negative definite. From Lyapunov theory we have
that $x_1(t+1)= [(I_M-\alpha D^{-1}L)- \ones[M] \ones[M]^\top]x_1(t)
$ is
globally uniformly asymptotically stable about the origin. From the
theory of symmetric quadratic forms we also have the following
inequality
\begin{equation}
\Delta V(x_1) \leq -(1-(1-\alpha \lambda_2^N)^2 )\|x_1\|^2,
\label{eqn:sc_proof_lyap1}
\end{equation}  
which gives us a convergence rate for the $x_1(t+1)=[(I_M-\alpha
D^{-1}L)- \ones[M] \ones[M]^\top]x_1(t)$ dynamics.
\oprocend
\end{pf}
  Now we will analyze the second subsystem. The Matrix $W$ has 1 as the simple eigenvalue with eigenvector
  $\ones[M]$. The matrix $\overline{W}=W-\frac{\ones[M]^\top
    \ones[M]}{M}$ is Schur stable and $\lambda^{\overline{W}} \in
  (-\frac{n-2}{n},1-\lambda_2^W)$, where $\lambda_2^W \in [0,1]$ is
  the second smallest eigenvalue associated with the weighted graph
  $\subscr{\mathcal{G}}{w}$. Hence we will analyze the convergence of the system $x_2(t+1)=\overline{W}x_2(t)$, which will give us the convergence rate for system $f_2$.


\begin{lemma}
  The system $x_2(t+1)=\overline{W}x_2(t)$ is
  globally uniformly asymptotically stable to the origin, and the convergence rate of system $f_2$ is proportional to $(1-(1-\lambda_2^W)^2 )$.
\label{lem:f_2_stab}
\end{lemma}
\begin{pf}
  With $\Delta V(x_2)= V (x_2(t+1)) - V (x_2(t))$ and
  $Q_2=\overline{W}^\top \overline{W} - I_M$ we have
  \begin{equation*}
    \Delta V(x_2) = x_2^\top(t)Q_2x_2(t) < 0 .
  \end{equation*}
  This follows from the fact that the eigenvalues $\lambda_{Q_2} \in
  ((\frac{n-2}{n})^2-1,(1-\lambda_2^W)^2-1)$, which makes $Q_2$
  negative definite. Hence according to Lyapunov theory
  $x_2(t+1) = f_2(x_2(t))-\frac{\ones[M]^\top \ones[M]}{M}x_2(t)$ is globally uniformly
  asymptotically stable  to the origin. In addition, 
  \begin{equation}
    \Delta V_2(x_2) \leq -(1-(1-\lambda_2^W)^2 )\|x_2\|^2,
    \label{eqn:sc_proof_lyap2}
  \end{equation}
  which finally gives us the convergence rate for the $f_2(x_2)$ 
  dynamics.
  \oprocend
\end{pf}

The analysis of the third subsystem, $x_3(t+1)=0$, is trivial.  Now, let
us define the following constants: $\delta_1=1-(1-\alpha
\lambda_2^N)^2 ,\ \ \delta_2=1-(1-\lambda_2^W)^2, \ \ \delta_3=1$, 
$\gamma_{11}=\gamma_{13}=\gamma_{22}=\gamma_{32}=\gamma_{33}=0$, 
$\gamma_{12}=k^p$, $ \gamma_{21}=\gamma_{23}=\gamma_{31}=1 ,
\beta_k=1,$ and $ \phi(x_k)=\|x_k\|,\; \forall k \in \{1,2,3 \}. $

Now we are ready to state the stability of System~\eqref{eqn:sc_ss}. 

\subsection*{Proof of Theorm \ref{thm:stab}}
\begin{pf}
  The system \eqref{eqn:sc_ss} can be equivalently represented in
  the form \eqref{eqn:sc_proof}. Now let us first consider driftless
  system \eqref{eqn:sc_proof}. The positive definite Lyapunov
  functions $V_k(x_k) \equiv V(x_k)$ and the interconnection functions
  $g_k(x)$ satisfy the conditions of \eqref{eqn:comp_lyap_cond} for
  all $t \geq 0$.
From Lemma~\ref{lem:f_1_stab} and Lemma~\ref{lem:f_2_stab} for each
subsystem $k,l \in \{1,2,3 \} $ we have:
\begin{subequations} \label{eqn:comp_lyap_cond}
\begin{align}
\Delta V(x_k) \leq & -\delta_k \phi^2(x_k), \\ 
\Big\| \frac{\partial V(x_k)}{\partial x_k} \Big\| \leq & \beta_k \phi(x_k), \\
\| g_k(t,x)\|  \leq & \sum_{l=1}^3 \gamma_{kl} \phi (x_k). 
\end{align}
\end{subequations}

Now if we consider a diagonal matrix $\diag{\delta} \in \real^{3
  \times 3}$ with
diagonal entries $(\delta_1,\delta_2,\delta_3)$, a column vector
$\beta=(\beta_1,\beta_2,\beta_3)^\top$ and a matrix $\Gamma=
(\gamma_{kl}) \in \real^{3\times 3} $, we can define a Matrix
   $S \in \real^{3 \times 3}$ as follows
\begin{equation}
S= \diag{\delta}- \beta \Gamma . 
\label{eqn:sc_proof_Smatrix}
\end{equation}
\begin{equation*}
    S=\begin{bmatrix}
      1-(1-\alpha \lambda_2^N)^2 & -k^p & 0 \\
      -1 & 1-(1-\lambda_2^W)^2 & -1 \\
      -1 & 0 & 1
    \end{bmatrix}.
\end{equation*}
The Matrix $S$ is an M-matrix, which is characterized by non-positive off diagonal entries and
positive leading principal minors. The former property is satisfied by
inspecting $S$. The first leading principal minor is positive from the
definition of the constants and the connectivity Assumption~1. For the
second leading principal minor to be positive we require $k^p <
\delta_1 \delta_2 $. For the third leading principal minor (det($S$))
to be positive we require $k^p < \frac{\delta_1
  \delta_2}{2} $.

  
   Now, we choose $k^p$ accordingly such that $S$ defined according to
  \eqref{eqn:sc_proof_Smatrix} is an M matrix. Now from
  \cite{HKK:02} (cf.~Theorem~9.2) we can conclude that the
  interconnected system~\eqref{eqn:sc_proof} is globally stable. 
  
  As the interconnections are asymptically stable we can infer the
following. Firstly, the subsystem~\eqref{eqn:sc_proof1} denoting the
position of the agents stabilizes to the desired shape, orientation
and scaling according to \cite{JC:08-acc} and also reach the desired
centroid due to the shifting term $k^p\ones[M]c_d$. Additionally, assumptions in executing the FODAC
algorithm in \cite{MZ-SM:08a} are satisfied due to the current assumption, and the fact that the first order differences of the reference signal are asymptotically stable from Lemma~\ref{lem:f_1_stab}. 
Thus, having satisfied all the required assumptions in executing the FODAC
algorithm in \cite{MZ-SM:08a}, 
   the centroid estimate $\hat{c}(t)$ converges to
$p(t)$. Thus, the overall system converges to the desired state
$X^d$. \oprocend
\end{pf}
  
\subsection*{Proof of Corollary \ref{corr:1}}

\begin{pf}
  For connected graphs we know that $\lambda_2^N \in
  (0,\frac{n}{n-1}]$ and $\lambda_2^W \in (0,1]$. From Equations~\eqref{eqn:sc_proof_lyap1} and~\eqref{eqn:sc_proof_lyap2}  we can see that the convergence of the
  swarm dynamics is faster with higher
  values of $\lambda_2^N$ and $\lambda_2^W$. 
  We know that the convergence rate for reaching the
  desired centroid is directly proportional to the control gain
  $k^p$. However from Theorm~\ref{thm:stab} we need to satisfy:
\begin{equation}
  k^p< \frac{1}{2}(1-(1-\alpha \lambda_2^N)^2)(1-(1-\lambda_2^W)^2)
\label{eqn:kp_bounds}
\end{equation}

Equation~\eqref{eqn:kp_bounds} reveals that, given a fixed $\alpha$,
the upperbound on $k^p$ can be increased with an increase in
$\lambda_2^W$ and $\lambda_2^N$.Hence, we have faster convergence of system~\eqref{eqn:sc_ss} with higher graph connectivity. 
\oprocend
\end{pf}  
                                                   








 \bibliography{alias,SM,JC,SMD-add}
\end{document}